\documentclass{article}

\usepackage{arxiv}

\usepackage[utf8]{inputenc} % allow utf-8 input
\usepackage[T1]{fontenc}    % use 8-bit T1 fonts
\usepackage{hyperref}       % hyperlinks
\usepackage{url}            % simple URL typesetting
\usepackage{booktabs}       % professional-quality tables
\usepackage{amsfonts}       % blackboard math symbols
\usepackage{nicefrac}       % compact symbols for 1/2, etc.
\usepackage{microtype}      % microtypography
\usepackage{lipsum}         % dummy text

\usepackage{graphicx}       % figure
\usepackage{subcaption}     % figure

\usepackage{verbatim}       % comment

\usepackage{cite}           % cite
\usepackage{amsmath}        % math

\usepackage{booktabs}       % table
\usepackage{multirow}       % table

\usepackage[ruled,vlined]{algorithm2e}

\usepackage{setspace}       % line spacing

\title{Learning Structural Graph Layouts and 3D Shapes for Long Span Bridges 3D Reconstruction}

\author{	
  Fangqiao Hu \\
    School of Civil Engineering \\
    Harbin Institute of Technology \\
    Harbin, China 150090 \\
  \texttt{hufangqiao@foxmail.com} \\
  \And  
  Jin Zhao \\
    School of Civil Engineering \\
    Harbin Institute of Technology \\
    Harbin, China 150090 \\
  \texttt{zhaojinha@hit.edu.cn} \\
  \And  
  Yong Huang \\
    School of Civil Engineering \\
    Harbin Institute of Technology \\
    Harbin, China 150090 \\
  \texttt{huangyong@hit.edu.cn} \\
  \And  
 Hui Li
 \thanks{Correspondence to: Dr. Hui Li (Professor), School of Civil Engineering, Harbin Institute of Technology, Harbin, China. Artificial Intelligence Lab, Harbin Institute of Technology, Harbin, China. E-mail: lihui@hit.edu.cn} \\
    School of Civil Engineering \\
    Harbin Institute of Technology \\
    Harbin, China 150090 \\
  \texttt{lihui@hit.edu.cn} \\
}

\begin{document}
\maketitle

\begin{abstract}
A learning-based 3D reconstruction method for long-span bridges is proposed in this paper. 3D reconstruction generates a 3D computer model of a real object or scene from images, it involves many stages and open problems. Existing point-based methods focus on generating 3D point clouds and their reconstructed polygonal mesh or fitting-based geometrical models in urban scenes civil structures reconstruction within Manhattan world constrains and have made great achievements. Difficulties arise when an attempt is made to transfer these systems to structures with complex topology and part relations like steel trusses and long-span bridges, this could be attributed to point clouds are often unevenly distributed with noise and suffer from occlusions and incompletion, recovering a satisfactory 3D model from these highly unstructured point clouds in a bottom-up pattern while preserving the geometrical and topological properties makes enormous challenge to existing algorithms. Considering the prior human knowledge that these structures are in conformity to regular spatial layouts in terms of components, a learning-based topology-aware 3D reconstruction method which can obtain high-level structural graph layouts and low-level 3D shapes from images is proposed in this paper. We demonstrate the feasibility of this method by testing on two real long-span steel truss cable-stayed bridges.
\end{abstract}


\section{INTRODUCTION}
The aim of this paper is to learn a 3D model of long-span bridges (hereinafter referred to as "bridges") from images captured by unmanned aerial vehicles (UAVs) fully automatically.

Recent developments in UAVs have heightened applications in wide range of industrial scenarios. UAVs can explore inaccessible areas carrying various types of sensors, such as digital cameras, infrared cameras, laser scanners (LiDAR) and so on, thus expected to play an increasingly important role in civil structure visual inspection systems. Cameras mounted on UAVs can capture images of the structure from various view points and record the exterior state of the structure. However, current visual inspection strategies often provide massive sets of unstructured and unaligned digital images, which take a lot of human efforts to filter and organize for further usage, the problem of how to fully automate this task becomes urgent that should be settled. One feasible solution is to integrate these images into a single 3D model of the target structure, which enables inspectors to engage with these images in a more intuitive manner and can provide a better monitoring scheme by recording and visualizing the life cycle of the entire structure. However, existing 3D models are as-designed models which somehow differ from current as-is models and failed to express the current exterior state of the structure, so recording or scanning the structure on the site is believed to be the only feasible solution.

Up to now, the research on civil structure 3D reconstruction has tended to focus on local points (i.e., pixels in image grids and points in 3D Euclidean space) rather than global features. These methods first obtain 3D coordinates of key-points and camera parameters from multiple images using structure-from-motion (SfM) \cite{hartley2003multiple,wu2011visualsfm,schonberger2016structure}, then generate dense 3D point clouds with multi-view stereo (MVS) \cite{furukawa2015multi,furukawa2010accurate}. 

Point clouds are rarely directly used in practice, since they are actually unstructured point sets and are not capable for texture mapping. Based on point clouds, there are two post processing, they are surface reconstruction \cite{delaunay1934sphere, kazhdan2006poisson, kazhdan2013screened} and fitting-based point cloud modeling methods \cite{furukawa2009manhattan, sinha2009piecewise, monszpart2015rapter, xiong2015flexible, holzmann2018semantically, raposo2018piecewise, kaiser2019survey, schnabel2007efficient, li2011globfit, rabbani2007integrated, attene2010hierarchical, li2018supervised, li2019primitive, li2016manhattan, li2016fitting, lee2013skeleton, patil2017adaptive, guo2016improved, ortega2019characterization, barazzetti2016parametric, dimitrov2016non, labatut2009hierarchical, lafarge2012creating, lafarge2013hybrid, lafarge2013surface, gorelick2017google}.

Surface reconstruction generates a polygonal mesh (a triangular mesh in most cases) as a surface of a 3D point cloud directly. A triangular mesh is represented as a set of $n$ ordered 3D vertices and $m$ triangular faces expressed as $\mathcal{M}=\{V\in \mathbb{R}^{n \times 3}, F\in \mathbb{N}_{+}^{m\times 3}\}$, each face use 3 vertices indexes. However, a polygonal mesh is often overly complex for civil structures in urban scenes where most surfaces are flat. Worse still, the meshing quality may degrade significantly because: (a) a MVS point cloud often suffers from severe noise and uneven distribution, which is different from LiDAR point clouds; (b) the target 3D object is often with complex geometry and topology, such as for a cable-stayed bridge in this paper; (c) no structure priors are introduced in surface reconstruction methods.

\begin{figure}[htbp]	
	\centering	
	\includegraphics[width=\textwidth]{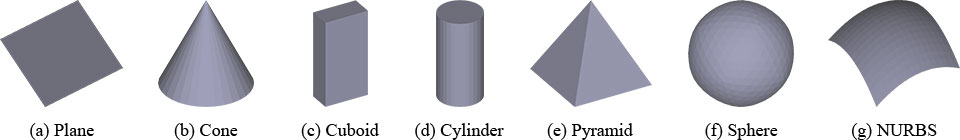}
	\caption{Basic geometric shapes.}
	\label{fig:primitives}
\end{figure}

Fitting-based modeling methods generate a compact geometric model to fit a 3D point cloud, where basic geometric shapes (e.g., planes, cones, cuboids, cylinders, pyramids, spheres, etc.) are employed, see Figure \ref{fig:primitives}. Sometimes point cloud segmentation algorithms (see \cite{grilli2017review} to get a literature review) are used for pre-processing. Fitting-based methods are summarized into: (a) piece-wise planes fitting; (b) 3D geometric primitives fitting; (c) non-uniform rational b-spline (NURBS) curves and surfaces fitting; (d) Hybrid fitting. They are briefly discussed in the following.

Piece-wise planar fitting methods use planes to fit a point cloud. Manhattan-world prior is introduced due to it is common in the real-world buildings \cite{furukawa2009manhattan, sinha2009piecewise}. Planar fitting has achieved remarkable results in building exterior reconstruction \cite{sinha2009piecewise, monszpart2015rapter, xiong2015flexible}, Holzmann et al. \cite{holzmann2018semantically} use lines in addition to point clouds, Raposo et al. \cite{raposo2018piecewise} use planes rather than being limited by point clouds, both achieving better results. 

3D geometric primitives fitting methods use 3D geometric primitives to fit a point cloud, sometimes constructive solid geometry (CSG) models are employed. Representing a 3D object with 3D geometric primitives is the most commonly used in computer vision and computer graphics fields \cite{kaiser2019survey}. There are many different methods to fit 3D geometric primitives including RANSAC \cite{schnabel2007efficient, li2011globfit, li2016manhattan}, Hough transform \cite{rabbani2007integrated}, primitive-driven region growth \cite{attene2010hierarchical}, learning-based methods \cite{li2018supervised, li2019primitive}, etc., which has been widely used in building exterior reconstruction \cite{li2016manhattan, li2016fitting}, building information modeling (BIM) such as pipes \cite{lee2013skeleton, patil2017adaptive} etc., and have achieved remarkable results.

NURBS fitting use NURBS curve and surfaces to fit a point cloud, which is useful in curved lines and surfaces including power lines \cite{guo2016improved, ortega2019characterization}, complex building exterior and curved pipes \cite{barazzetti2016parametric, dimitrov2016non}, etc.

Hybrid fitting \cite{labatut2009hierarchical, lafarge2012creating, lafarge2013hybrid, lafarge2013surface} is employed as a technique solution using LiDAR point clouds, MVS point clouds, surface reconstruction, geometric primitives fitting, etc. For example, the Google Earth \cite{gorelick2017google}.

However, fitting-based methods failed to reconstruct the MVS point cloud in this paper due to: (a) noise, uneven distribution, missing points and occlusion in MVS point cloud; (b) high level structure priors are not introduced, since fitting-based methods only consider bottom-up fitting.

In summary, surface reconstruction and point cloud modeling are both susceptible to noise, they are in a bottom-up fashion, where high level structure priors are rarely introduced, this brings enormous challenge to existing algorithms.

Machine learning has long been a question of great interest in a wide range of fields including computer vision and computer graphics. In contrast to traditional point-based algorithms, which can be considered as a point-wise stereo vision measurement and does not rely on global feature and prior knowledge, learning-based methods \cite{jack2018learning, wang2018pixel2mesh, groueix2018papier, kanazawa2018learning, smith2018multi, tulsiani2017learning, zou20173d, niu2018im2struct, fan2017point, lun20173d, lin2018learning, choy20163d, yan2016perspective, kar2017learning, sun2018pix3d} learn global features from images. In general, these algorithms first encode single or multiple images into a latent vector by convolutional neural networks (CNNs) and then decode it into 3D models represented by: (a) polygonal mesh models \cite{jack2018learning, wang2018pixel2mesh, groueix2018papier, kanazawa2018learning, smith2018multi}; (b) geometric primitives models \cite{tulsiani2017learning, zou20173d, niu2018im2struct}; (c) point cloud models \cite{fan2017point, lun20173d, lin2018learning}; (d) volumetric models \cite{choy20163d, yan2016perspective, kar2017learning, sun2018pix3d}, in terms of output 3D representation forms. These methods work well in synthesized benchmark datasets \cite{xiang2014beyond, chang2015shapenet} but are still far from meeting the industrial engineering requirements. 

\begin{figure}[htbp]	
	\centering	
	\includegraphics[width=\textwidth]{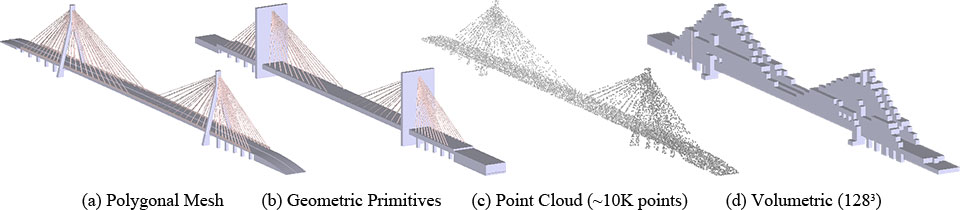}
	\caption{A comparison of different representation forms.}
	\label{fig:comparison}	
\end{figure}

Table \ref{tab:comparison} and Figure \ref{fig:comparison} shows a comparison of different representation forms of a bridge model, all these models are directly converted from the same original 3D bridge model. They are: (a) Triangular mesh, defined by a set $\{V\in \mathbb{R}^{n\times 3},F\in \mathbb{N}_{+}^{m\times 3}\}$ of vertices $V$ and faces $F$; (b) Geometric primitives, defined by a set $\{\{{B}_{i}\in \mathbb{R}^{8}\}_{i=1}^{m},\{{C}_{j}\in \mathbb{R}^{7}\}_{j=1}^{n}\}$ of $m$ cuboids and $n$ cylinders; (c) Point cloud, defined by a set $\{{P}_{i}\in \mathbb{R}^{3}\}_{i=1}^{n}$ of $n$ points; and (d) Volumetric model, defined by a 3D voxel grid $V \in \mathbb{R}^{n\times n\times n}$.

\begin{table}
	\centering
	\caption{A comparison of different representation forms.}
	\label{tab:comparison}
	\setlength\heavyrulewidth{0.35ex}
	\resizebox{\textwidth}{!}{%
	\begin{tabular}{llll} 
		\toprule
		Representation       & Expression                                                                             & Pros.                                                                                  & Cons.                                                                                           \\ 
		\midrule
		Triangular mesh      & $\{V\in \mathbb{R}^{n\times 3},F\in \mathbb{N}_{+}^{m\times 3}\}$                      & - High representation ability for details                                              & - Vertices connections are very hard to learn                                                   \\
		Geometric primitives & $\{\{{B}_{i}\in \mathbb{R}^{8}\}_{i=1}^{m},\{{C}_{j}\in \mathbb{R}^{7}\}_{j=1}^{n}\}$  & \begin{tabular}[c]{@{}l@{}}- Compact\\ - Parameterized\\ - Able to learn \end{tabular} & - Lose details                                                                                  \\
		Point cloud          & $\{{P}_{i}\in \mathbb{R}^{3}\}_{i=1}^{n}$                                              & \begin{tabular}[c]{@{}l@{}}- Manipulation flexibility \\ - Easy to learn \end{tabular} & \begin{tabular}[c]{@{}l@{}}- Low representation ability\\ - High memory overhead \end{tabular}  \\
		Volumetric model     & $V \in \mathbb{R}^{n\times n\times n}$                                                 & - Easy to learn                                                                        & \begin{tabular}[c]{@{}l@{}}- Low resolution\\ - High memory overhead \end{tabular}              \\
		\bottomrule
	\end{tabular}
	}
\end{table}

The objective of this paper is to learn a 3D model for bridges from UAV images. We use hybrid models including geometric primitives and volumetric models to represent a bridge object, and hybrid input data including images and point clouds.

\section{METHODOLOGY}
\begin{figure}[htbp]	
	\centering	
	\includegraphics[width=\textwidth]{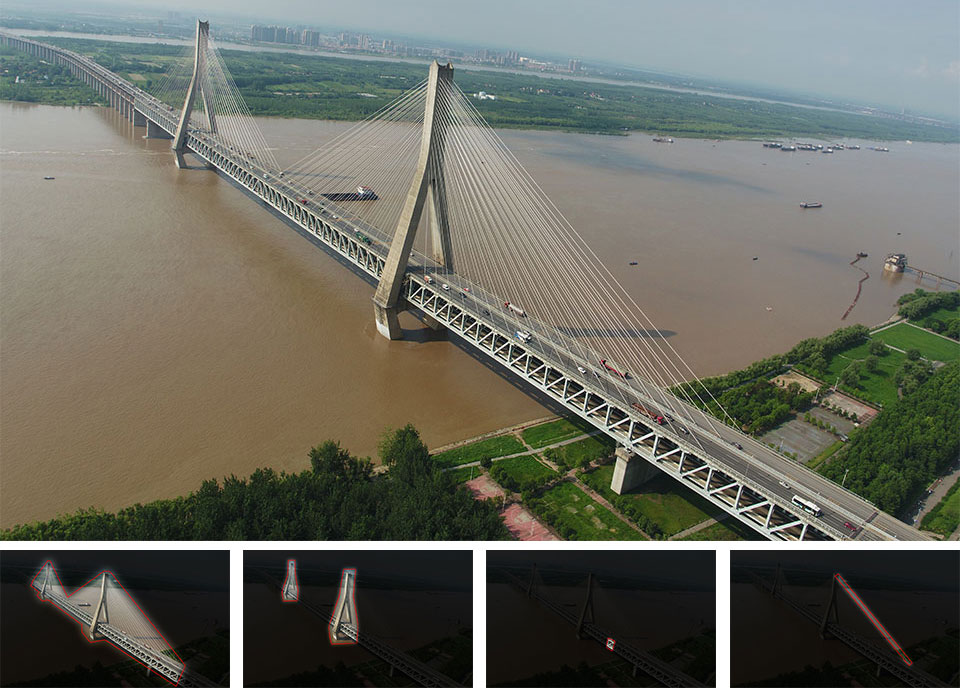}
	\caption{A picture of long-span cable-stayed bridge. Humans can delicately distinguish the bridge object in the picture from the background, then have information that this bridge has two towers arranged symmetrically, the truss blocks are aligned repeatedly and the cables are similar from one to another.}
	\label{fig:bridge}  
\end{figure}

\begin{figure}[htbp]	
	\centering	
	\includegraphics[width=\textwidth]{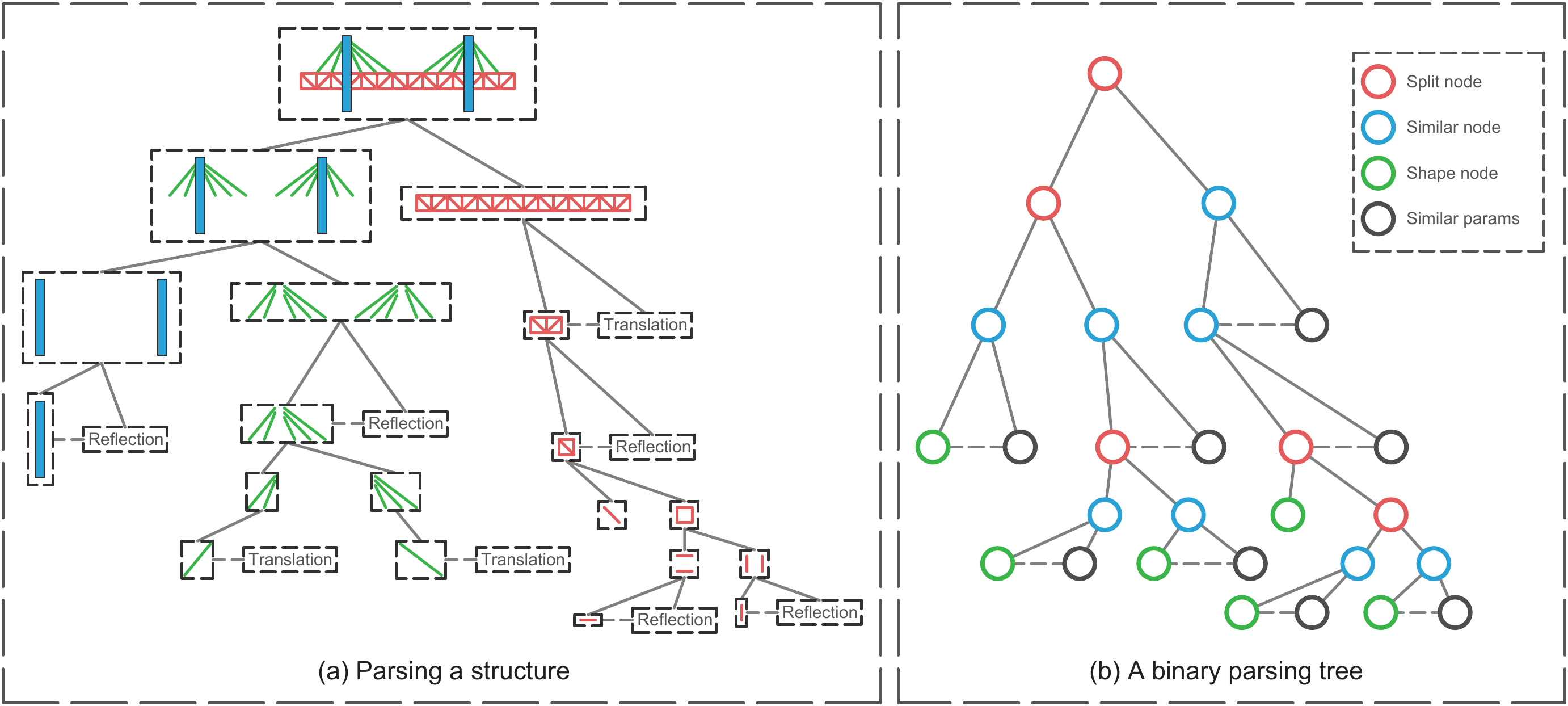}
	\caption{The parsing tree for a representative simplified cable-stayed bridge. (a) Structural graph layouts used for parsing this bridge, and (b) Corresponding hierarchical binary parsing tree (four node types are used in total).}
	\label{fig:binary_tree}  
\end{figure}

Only images of long-span bridges captured by UAVs are used for 3D reconstruction of bridges. Taking an example of seeing a picture of a bridge shown in Figure \ref{fig:bridge} by human to understand the bridge, one can delicately distinguish the bridge object in the picture from the background, then have further information that this bridge has two towers arranged symmetrically, the truss blocks are aligned repeatedly and the cables are similar from one to another. Inspired by the human recognition process on an object above, a hierarchical binary parsing tree is proposed to parse a bridge structure in this paper, as shown in Figure \ref{fig:binary_tree}.

The entire procedure for 3D reconstruction of a bridge is summarized as: (a) first, a 3D point cloud and camera parameters from images are obtained as 3D global initial information through a general rigid SfM and MVS pipeline; (b) then, regions of interest (RoIs) in 3D point cloud and 2D image are obtained to reduce the interference of background and noise; (c) following feeding 3D point cloud, images, and RoIs into the proposed learning framework, a binary tree to describe the structural graph layouts is determined via learning algorithm; (d) when the binary tree is obtained, 3D shapes are further learned from shape nodes; and (e) the final 3D model is assembled with structural graph layouts and 3D shapes.

\subsection{3D Point Cloud and Camera Parameters}
\label{sec:2_1}
\begin{figure}[htbp]	
	\centering	
	\includegraphics[width=\textwidth]{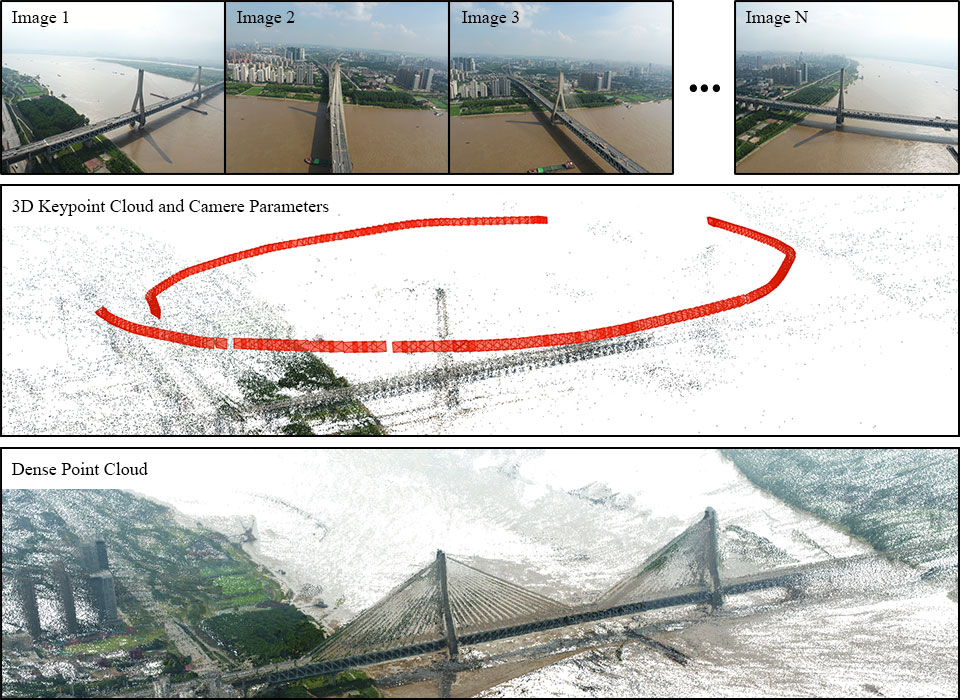}
	\caption{3D point cloud and camera parameters obtained by rigid SfM and MVS pipeline provided by \cite{schonberger2016structure}. Point clouds are represented by colored points, camera poses (i.e., locations and orientations) are represented by red blocks.}
	\label{fig:sfm_mvs}  
\end{figure}

This step is to obtain 3D point cloud and camera parameters (see Figure \ref{fig:sfm_mvs}) from images as 3D global initial information through a general rigid SfM and MVS pipeline. SfM obtains 3D coordinates of key-points and camera parameters from multiple images. It can be summarized as (a) 2D key-points extraction and matching; (b) camera parameters verification and (c) sparse 3D key-points reconstruction. MVS then takes the output (i.e., camera parameters and coordinators of 3D key-points) of SfM as input, and generates depth-maps, dense 3D point clouds and so on. These steps will be described in detail as follows.

\subsubsection{2D key-points extraction and matching}
\label{sec:2_1_1}
This step is to extract key-points and find key-point pairs in multi-view images. A 2D key-point is a coordinate in an image indicating that the pixel values in its neighboring pixels vary greatly. 

Given an image  , the scale invariant feature transform (SIFT) \cite{lowe2004distinctive} key-point extraction processing is formulated as
\begin{equation}
\begin{split}
L(x, y, \sigma) & =G(x, y, \sigma) \star I(x, y) \\
G(x, y, \sigma) & =\frac{1}{2 \pi \sigma^{2}} \exp \left(-\left(x^{2}+y^{2}\right) / 2 \sigma^{2}\right) \\
D(x, y, \sigma) & =L(x, y, \sigma)-L(x, y, k \sigma)
\end{split}
\end{equation}
where $(x,y)$ is pixel coordinate within an image, $L(x,y,\sigma )$ is a Gaussian blurred image parameterized by a factor $\sigma $ and generated by convoluting a Gaussian kernel $G(x,y,\sigma )$ with an input image $I(x,y)$. A pyramid-like scale space is constructed in \cite{lowe2004distinctive} to store Gaussian blurred images with different factors and different sizes to enable multi-scale detection. The factors are ($\sigma $, $k\sigma $, ${k}^{2}\sigma $, ${k}^{3}\sigma$, ...) in each octave, where $k={2}^{1/s}$, and $s$ is the number of intervals in each octave. The image size is halved from a lower octave to a higher one. $D(x,y,\sigma)$ is a difference-of-Gaussian image produced by subtracting two adjacent Gaussian blurred images in each octave. The key-point candidates are extracted by comparing each pixel in $D(x,y,\sigma)$ to its neighbors in the current factor and two adjacent factors in each octave, and is chosen if it is larger or smaller than all other neighboring pixels.

A key-point descriptor represented by a 128-D vector is needed to characterize a key-point. It assigns gradient magnitudes and orientations in a neighboring $16\times 16$ pixel grids region, each magnitude $m(x,y)$ and orientation $\theta (x,y)$ are expressed as
\begin{equation}
\begin{split}
m(x, y) & =\sqrt{(L(x+1, y)-L(x-1, y))^{2}+(L(x, y+1)-L(x, y-1))^{2}} \\
\theta(x, y) & =\arctan ((L(x, y+1)-L(x, y-1)) / (L(x+1, y)-L(x-1, y)))
\end{split}
\end{equation}
These gradient magnitudes and orientations are then accumulated into orientation histograms by summarizing the contents over $4\times 4$ subregions, which generates a descriptor in form of a $4\times 4$ array of histograms, each with 8 orientation bins, this results in a $4\times 4\times 8=128$ dimensional vector, which is then normalized to form the key-point descriptor.

Matching is to find all descriptor pairs in an image pair. Approximate nearest neighbors (ANN) \cite{muja2014scalable} is used to reduce algorithm complexity compared to exhaustive descriptor pairs matching. Given an image pair, for a query descriptor ${D}_{i}$ in one image, ANN finds the approximate closest descriptor ${D}_{j}$ (i.e., ${D}_{i} \cong {D}_{j}$) in the other image by constructing a priority search k-means tree w.r.t metric distance (here Euclidean distance $d({D}_{i},{D}_{j})=\left\| {{D}_{i}}-{{D}_{j}} \right\|_{2}$). 

In summary, given $N$ images, the SIFT detector extracts 2D key-points and their descriptors for all images, this results in a key-point set $\mathcal{K}=\left\{\left(p_{i, j}, D_{i, j}\right) | j=1, \ldots, M_{i}, i=1, \ldots, N\right\}$, where $p_{i, j} \in \mathbb{Z}^{2}$ denotes the $j$-th key-point in the $i$-th image and $D_{i, j} \in \mathbb{R}^{128}$ (a 128-D vector in \cite{lowe2004distinctive}) is the corresponding descriptor. These key-points are then matched via distances of descriptors for all image pairs, and this results in a set of matched key-point pairs $\left\{\left(p_{i, j}, D_{i, j}, p_{k, l}, D_{k j}\right) \in \mathcal{K} \times \mathcal{K} | D_{i, j} \cong D_{k, l}, i<k\right\}$. 

\subsubsection{Camera parameters verification}
\label{sec:2_1_2}
This step is to verify camera parameters for multiple views including intrinsic parameters (e.g., focal length) and extrinsic parameters (camera motions). In this section, homogeneous coordinate system is introduced to simplify calculations, e.g., a 2D key-point $(x,y)$ is expressed as $(x,y,1)$ in homogeneous coordinate, and subject to $\lambda (x,y,1)=(x,y,1)$.  In single view geometry, camera parameters $C\in \mathbb{R}^{3\times 4}$ defines a projective mapping $p=CP$, where $P\in \mathbb{R}^{4}$ and $p\in \mathbb{R}^{3}$ are two points in 3D world coordinate and 2D pixel coordinate in homogeneous coordinate system. Camera parameters $C$ includes intrinsic parameters $K\in \mathbb{R}^{3\times 3}$  which defines a projective mapping from 3D camera coordinate (cartesian coordinate system) to 2D pixel coordinate (homogeneous coordinate system), and extrinsic parameters $[R|t]\in {{\mathbb{R}}^{3\times 4}}$ where $R\in \mathbb{R}^{3\times 3}$ is a rotation matrix and $t\in \mathbb{R}^{3}$ is a translation vector, defining rotation and translation from 3D world coordinate (homogeneous coordinate system) to 3D camera coordinate (cartesian coordinate system). Camera lens radial distortion matrix $S_{\lambda}=diag(1/\lambda ,1/\lambda ,1)$ is introduced to eliminate the image distortion caused by camera lens, where $\lambda=1+\sum\limits_{p=1}^{3}{k}_{p}{d}^{2p}$ is a nonlinear polynomial function of the distance $d$ from the current pixel to image center, ${k}_{p}\in \mathbb{R}$ are radial distortion parameters. As a result, the camera parameters are defined by a matrix $C={{S}_{\lambda }}K[R|t]$.

In multi-view geometry, $K$ and $S_{\lambda}$ are shared among different views if they are obtained with the same camera. Note that $K$ and $S_{\lambda}$ are known, calibrated or estimated approximately beforehand and optimized at the last stage. Images are then undistorted with $S_{\lambda}$ to avert nonlinear mapping functions, i.e., only $C=K[R|t]$ is considered, hence only camera motion $[R|t]$ for each view is need to be solved. Generally, the camera coordinate of the first view is used as the world coordinate. In this study, $K$ and $S_{\lambda}$ are estimated from exchangeable image file format (EXIF) tags and are shared among different views.

To solve camera motion $[R|t]$, consider an image pair (i.e., the $i$-th and $j$-th image) in epipolar geometry (i.e., two views geometry), the camera parameters of two views are encapsulated in the fundamental matrix $F=K^{-T}[t]_{ \times} R K^{-1}$, where $t$ and $R$ describes the relative translation and rotation of two views, $[t]_{ \times}$ is a skew-symmetric matrix of $t$ for matrixes cross product. The fundamental matrix $F$ gives constraints on how the scene changes under two views, or how the coordinates of a matched point pair changes within an image pair. A matched key-point pair $p_{i}, p_{j} \in \mathbb{R}^{3}$ in homogeneous coordinate system fulfills $p_{\mathrm{i}}^{T} F p_{j}=0$, where $p_i$ and $p_j$ are two key-points in the $i$-th and $j$-th image. The first step is to solve $F$ subject to $rank(F)=2$ from key-point pairs, the 9 elements in $F$ can be solved by least square algorithm since an image pair has a large number of matched key-point pairs. The next step is to solve $R$ and $t$ from $F$, assume that the camera matrixes of the first and second view are $K[I|0]$ and $K[R|t]$ (i.e., the world coordinate is on the first camera coordinate), $R$ and $t$ can then be retrieved from $E={K}^{T}FK$  as following, suppose that the singular value decomposition (SVD) of $E$ is $E=Udiag(1,1,0){V}^{T}$ since $E$ has two equal singular values (refer to \cite{hartley2003multiple} for proof), the solution of $R$ and $t$ are up to a scale and a four-fold ambiguity, they are $t=u_3$ or $t=-u_3$ and $R=UWV^T$ or $R=U{W}^{T}{V}^{T}$, where 
$W=\left[ \begin{matrix}
0 & -1 & 0  \\
1 & 0 & 0  \\
0 & 0 & 1  \\
\end{matrix} \right]$
 is orthogonal and skew-symmetric. A solution of $R$ and $t$ is valid only when a reconstructed point is in front of both cameras.

Random sample consensus (RANSAC) \cite{fischler1981random} is an iterative method to remove outliers in key-points matching and improve epipolar geometry accuracy, where the epipolar geometry is estimated by randomly and iteratively sampled key-point pairs rather than by all key-point pairs.

\subsubsection{3D Key-points reconstruction}
\label{sec:2_1_3}
This step is to reconstruct 3D key-points from 2D key-point pairs. A 3D point $P\in \mathbb{R}^4$ in homogeneous representation is obtained by triangulation algorithm $P \sim \arg \min _{P}\left(\left\|p_{i}-C_{i} P\right\|_{2}+\left\|p_{j}-C_{j} P\right\|_{2}\right)$ for each matched key-point pair $p_i$ and $p_j$ via camera matrixes $C_i$ and $C_j$ in the $i$-th and $j$-th image, the $\sim$ sign implies that there is a non-zero scale factor since homogeneous representation is involved. 

For multiple views, Incremental Structure-from-Motion \cite{wu2011visualsfm, schonberger2016structure} first initializes with a two-view reconstruction, then registers other views to the current reconstruction one-by-one. A new view image is registered to the current reconstruction if it observes existing 3D points, i.e., the key-points in the new view has an overlap to the current views w.r.t key-point descriptors. Camera motion of the new view is then estimated with Perspective-n-Point algorithm from $n$ corresponding 3D and 2D points, expressed as $[R | t] \cong \arg \min _{R, t} \sum_{i=1}^{n}\left\|p_{i}-K[R | t] P_{i}\right\|_{2}$, where $\left\{P_{i}\right\}_{i=1}^{n}$ and $\left\{p_{i}\right\}_{i=1}^{n}$ are corresponding 3D and 2D points w.r.t key-point descriptors. New 3D points are then obtained with triangulation algorithm in a pair-by-pair mode and results in a new reconstruction. The reconstruction is completed if no new image can be registered. To record multi-view correspondence of 2D points among multiple views, a track is defined for a reconstructed 3D point $P_i$, i.e., a list of corresponding (w.r.t key-point descriptors) 2D points $\left(p_{1, i}, p_{2, i}, \ldots\right)$ for different new views.

Bundle adjustment refines all camera parameters $\left\{C_{i} \in \mathbb{R}^{3 \times 4}\right\}_{i=1}^{N}$ for $N$ views and positions of all $M$ reconstructed 3D key-points $\left\{P_{j} \in \mathbb{R}^{4}\right\}_{j=1}^{M}$ to minimize the overall re-projection error, expressed as $E=\sum_{i=1}^{N} \sum_{j=1}^{M}\left\|p_{i, j}-C_{i} P_{j}\right\|_{2}$ after every incremental reconstruction.

\subsubsection{Multi-view stereo (MVS)}
\label{sec:2_1_4}
This step is to generate dense point cloud. The output of SfM are sparse 3D key-points and camera parameters, to get dense 3D points, the depth value for all pixels need to be calculated via epipolar geometry. 

Given an image pair and one pixel $p_i$ within the first image, the preimage of $p_i$ is a ray that goes from the camera center and across $p_i$. To find the corresponding pixel $p_j$ in the second image, consider epipolar geometry, all $p_j$ that fulfill the epipolar geometry $p_{j}^{T} F^{T} p_{i}=0$ make up the epipolar line $l_{j}=F^{T} p_{i}$ and have $p_{j}^{T} l_{j}=0$. In fact, $p_j$ should lie on the projected line of the preimage on the second image, i.e., $p_j$ lies on the epipolar line $l_j$. The search for the corresponding pixel $p_j$ is to find the best-matched pixel to $p_i$ along the epipolar line $l_j$, this processing is done via a sliding window rather than a single pixel. Consider a fixed window around the point $p_i$ in the first image and a slide window along the epipolar line $l_j$ in the second image, Normalized Cross Correlation (NCC) compares the score of the pixel values within the two windows, expressed as $c=\frac{\left(w_{i}-\overline{w_{i}}\right) \cdot\left(w_{j}-\overline{w_{j}}\right)}{\left\|\left(w_{i}-\overline{w_{i}}\right)\right\|_{2}\left\|\left(w_{j}-\overline{w_{j}}\right)\right\|_{2}}$, where $w_i$ and $w_j$ are vectorized pixel values within the first and second window, $\overline{w_{i}}$ and $\overline{w_{j}}$ are mean values, and $\cdot$ denotes inner product. The corresponding pixel is selected in the second image when NCC score $c$ achieves maximum. The depth values of two corresponding pixels are then calculated by triangulation algorithm mentioned in Section 2.1.3, that is, the depth value for $p_i$ is the distance from the first camera center to its reconstructed 3D point $P$, the same with $p_j$.

By calculating all pixel depth values within an image, a depth-map $\mathcal{I} \in \mathbb{R}^{h \times w}$ is obtained, which shares the identical camera parameters $C$ with its original RGB image. By treating a depth-map as a 2D array of 3D points, multiple depth-maps can be considered as a merged 3D point cloud. Patch-based multi-view stereo (PMVS) \cite{furukawa2010accurate} is an alternative algorithm to generate dense point cloud. Figure \ref{fig:sfm_mvs} shows 3D key-points, dense point cloud and camera poses.

\subsection{Region of interest (RoI) in 3D and 2D}
\label{sec:2_2}
This step is to obtain region of interest (RoI) in 3D point cloud and 2D images. RoI is necessary since background and noise interfere algorithms significantly \cite{georgakis2017synthesizing}.

\subsubsection{RoI in 3D: a 3D orientated bounding box (3D OBB)}
\label{sec:2_2_1}
3D orientated bounding box (3D OBB) is an approximate solution to find RoI in 3D point cloud. The raw point cloud is with a large range including many environmental points, we first filter a point cloud with criterion below
\begin{equation}
P^{out}=\{P_{i} | \left\|P_{i}-\mu\right\|_{2} < k \sigma, P_{i} \in P^{in} \}
\end{equation}
where $P^{in}$ and $P^{out}$ denote the input and output 3D point sets, $\mu$ and $\sigma$ stand for the mean and standard deviation of $P^{in}$, $k$ is a fixed scale factor. This results in a sphere boundary parameterized by center $\mu$ and radius $k\sigma$. $k=1$ works well in our experiments.

A 3D OBB has 9 degrees of freedom (DoFs) including a 3D translation $(t_{x}, t_{y}, t_{z})$ indicating OBB center coordinates, 3 rotation angles $(r_{x}, r_{y}, r_{z})$ and 3 dimensional parameters including length, width and height expressed as $(d_{x}, d_{y}, d_{z})$. The baseline of camera is rectified to horizontal in advance, if not, many image processing tools can rectify horizon line to horizontal by rotating images. So only one rotation angle $r_z$ instead of $(r_{x}, r_{y}, r_{z})$ is needed to determine the 3D OBB, hence a 7-DoFs 3D OBB is denoted as $B=(t_{x}, t_{y}, r_{z}, d_{x}, d_{y}, t_{z}, d_{z})$.

\begin{figure}[htbp]	
	\centering	
	\includegraphics[width=\textwidth]{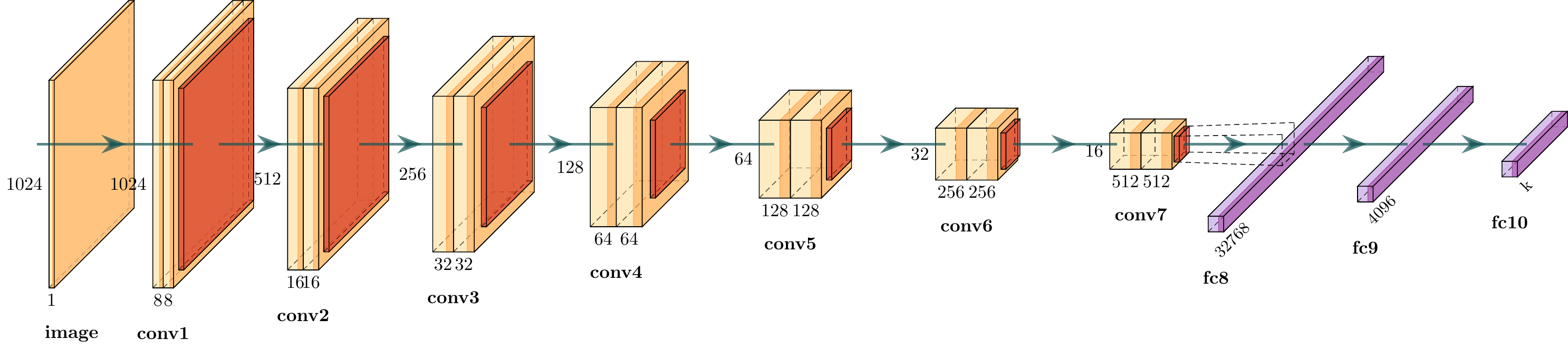}
	\caption{The proposed CNN. the input of this network is a $1024\times 1024\times 1$ dimensional image, the output is a $k$ dimensional vector. From conv1 to conv7, each of them contains 2 convolutional (Conv) layers, 2 batch normalization (BN) layers, 2 activation (PReLU) layers and 1 max pooling (MP) layer, composed in sequence of Conv – BN – PReLU – Conv – BN – PReLU – MP. For the last 3 fully connected (FC) layers from fc8 to fc10, each of them is composed in sequence of FC – PReLU.}
	\label{fig:3dobb_cnn}
\end{figure}

\begin{figure}[htbp]	
	\centering	
	\includegraphics[width=\textwidth]{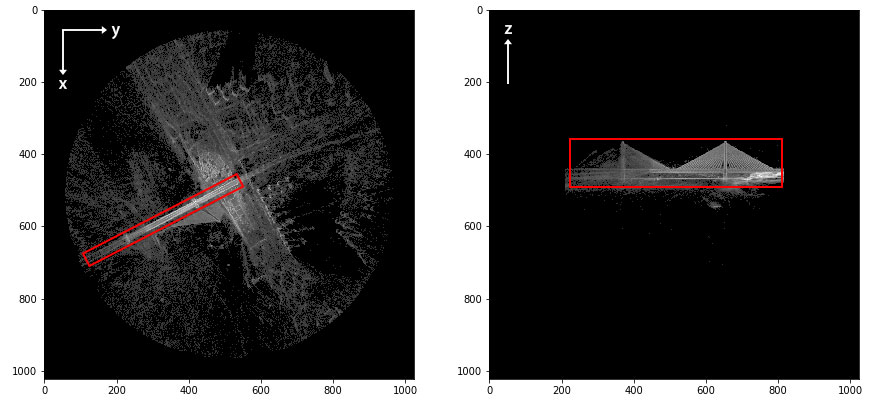}
	\caption{The obtained 3D OBB $B=(t_{x}, t_{y}, r_{z}, d_{x}, d_{y}, t_{z}, d_{z})$. In the first-step CNN, a 2D OBB $(t_{x}, t_{y}, r_{z}, d_{x}, d_{y})$ is obtained from the bird view (left). Known 2D OBB in the bird view, the second-step CNN is to get $(t_{z}, d_{z})$ from the front view of the bridge (right).}
	\label{fig:3dobb}
\end{figure}

We design a simple yet efficient two-step convolutional neural network to obtain the 7-DoFs 3D OBB. The configuration of the CNN (see Figure \ref{fig:3dobb_cnn}) follows VGG \cite{simonyan2014very} style. In Figure \ref{fig:3dobb_cnn}, the input of this network is a $1024\times 1024\times 1$ dimensional image, the output is a $k$ dimensional vector. In the first step, the input is a gray scale bird view image, the output is a 2D OBB $\left(t_{x}, t_{y}, r_{z}, d_{x}, d_{y}\right)$ with dimension $k=5$. On applying the 2D OBB to filter and align the point cloud, we can get the front view image. In the second step, the input is a gray scale front view image, the output is $\left(t_{z}, d_{z}\right)$ with dimension $k=2$. After that, the 3D point cloud is filtered with the 7-DoFs OBB. see Figure \ref{fig:3dobb}. Table \ref{tab:3dobb_cnn} lists the detailed properties of layers and operators. We will describe the function of these layers in following.

\begin{table}[htbp]
	\centering
	\caption{Detailed properties of layers and operators.}
	\label{tab:3dobb_cnn}
	\setlength\heavyrulewidth{0.35ex}
	\def\arraystretch{0.95}%  1 is the default, change whatever you need
	\begin{tabular}{lllllll}
		\toprule
		Layers & Feature size                & Operators     & Kernel size             & No. & Stride & Padding \\
		\midrule
		0      & $1024 \times 1024 \times 1$ & Input         & -                       & -   & -      & -       \\
		1      & $1024 \times 1024 \times 8$ & Conv          & $3 \times 3 \times 1$   & 8   & 1      & 1       \\
		2      & $1024 \times 1024 \times 8$ & BN            & -                       & -   & -      & -       \\
		3      & $1024 \times 1024 \times 8$ & PReLU         & -                       & -   & -      & -       \\
		4      & $1024 \times 1024 \times 8$ & Conv          & $3 \times 3 \times 8$   & 8   & 1      & 1       \\
		5      & $1024 \times 1024 \times 8$ & BN            & -                       & -   & -      & -       \\
		6      & $1024 \times 1024 \times 8$ & PReLU         & -                       & -   & -      & -       \\
		7      & $512 \times 512 \times 8$   & MP            & $2 \times 2 \times 8$   & -   & 2      & -       \\
		8      & $512 \times 512 \times 16$  & Conv          & $3 \times 3 \times 8$   & 16  & 1      & 1       \\
		9      & $512 \times 512 \times 16$  & BN            & -                       & -   & -      & -       \\
		10     & $512 \times 512 \times 16$  & PReLU         & -                       & -   & -      & -       \\
		11     & $512 \times 512 \times 16$  & Conv          & $3 \times 3 \times 16$  & 16  & 1      & 1       \\
		12     & $512 \times 512 \times 16$  & BN            & -                       & -   & -      & -       \\
		13     & $512 \times 512 \times 16$  & PReLU         & -                       & -   & -      & -       \\
		14     & $256 \times 256 \times 16$  & MP            & $2 \times 2 \times 16$  & -   & 2      & -       \\
		15     & $256 \times 256 \times 32$  & Conv          & $3 \times 3 \times 16$  & 32  & 1      & 1       \\
		16     & $256 \times 256 \times 32$  & BN            & -                       & -   & -      & -       \\
		17     & $256 \times 256 \times 32$  & PReLU         & -                       & -   & -      & -       \\
		18     & $256 \times 256 \times 32$  & Conv          & $3 \times 3 \times 32$  & 32  & 1      & 1       \\
		19     & $256 \times 256 \times 32$  & BN            & -                       & -   & -      & -       \\
		20     & $256 \times 256 \times 32$  & PReLU         & -                       & -   & -      & -       \\
		21     & $128 \times 128 \times 32$  & MP            & $2 \times 2 \times 32$  & -   & 2      & -       \\
		22     & $128 \times 128 \times 64$  & Conv          & $3 \times 3 \times 32$  & 64  & 1      & 1       \\
		23     & $128 \times 128 \times 64$  & BN            & -                       & -   & -      & -       \\
		24     & $128 \times 128 \times 64$  & PReLU         & -                       & -   & -      & -       \\
		25     & $128 \times 128 \times 64$  & Conv          & $3 \times 3 \times 64$  & 64  & 1      & 1       \\
		26     & $128 \times 128 \times 64$  & BN            & -                       & -   & -      & -       \\
		27     & $128 \times 128 \times 64$  & PReLU         & -                       & -   & -      & -       \\
		28     & $64 \times 64 \times 64$    & MP            & $2 \times 2 \times 64$  & -   & 2      & -       \\
		29     & $64 \times 64 \times 128$   & Conv          & $3 \times 3 \times 64$  & 128 & 1      & 1       \\
		30     & $64 \times 64 \times 128$   & BN            & -                       & -   & -      & -       \\
		31     & $64 \times 64 \times 128$   & PReLU         & -                       & -   & -      & -       \\
		32     & $64 \times 64 \times 128$   & Conv          & $3 \times 3 \times 128$ & 128 & 1      & 1       \\
		33     & $64 \times 64 \times 128$   & BN            & -                       & -   & -      & -       \\
		34     & $64 \times 64 \times 128$   & PReLU         & -                       & -   & -      & -       \\
		35     & $32 \times 32 \times 128$   & MP            & $2 \times 2 \times 128$ & -   & 2      & -       \\
		36     & $32 \times 32 \times 256$   & Conv          & $3 \times 3 \times 128$ & 256 & 1      & 1       \\
		37     & $32 \times 32 \times 256$   & BN            & -                       & -   & -      & -       \\
		38     & $32 \times 32 \times 256$   & PReLU         & -                       & -   & -      & -       \\
		39     & $32 \times 32 \times 256$   & Conv          & $3 \times 3 \times 256$ & 256 & 1      & 1       \\
		40     & $32 \times 32 \times 256$   & BN            & -                       & -   & -      & -       \\
		41     & $32 \times 32 \times 256$   & PReLU         & -                       & -   & -      & -       \\
		42     & $16 \times 16 \times 256$   & MP            & $2 \times 2 \times 256$ & -   & 2      & -       \\
		43     & $16 \times 16 \times 512$   & Conv          & $3 \times 3 \times 256$ & 512 & 1      & 1       \\
		44     & $16 \times 16 \times 512$   & BN            & -                       & -   & -      & -       \\
		45     & $16 \times 16 \times 512$   & PReLU         & -                       & -   & -      & -       \\
		46     & $16 \times 16 \times 512$   & Conv          & $3 \times 3 \times 512$ & 512 & 1      & 1       \\
		47     & $16 \times 16 \times 512$   & BN            & -                       & -   & -      & -       \\
		48     & $16 \times 16 \times 512$   & PReLU         & -                       & -   & -      & -       \\
		49     & $8 \times 8 \times 512$     & MP            & $2 \times 2 \times 512$ & -   & 2      & -       \\
		50     & $32768$                     & Re-shape      & -                       & -   & -      & -       \\
		51     & $4096$                      & FC            & $32768 \times 4096$     & -   & -      & -       \\
		52     & $4096$                      & Dropout (0.5) & -                       & -   & -      & -       \\
		53     & $4096$                      & BN            & -                       & -   & -      & -       \\
		54     & $4096$                      & PReLU         & -                       & -   & -      & -       \\
		55     & $k$                         & FC            & $4096 \times k$         & -   & -      & -       \\
		56     & $k$                         & Output        & -                       & -   & -      & -       \\
		\bottomrule
	\end{tabular}%
\end{table}

Convolutional operation extract features from an image through a convolutional kernel. One single channel of an image is viewed as a matrix, so convolution indicates 2-dimensional discrete convolution, expressed as
\begin{equation}
I_{out}[x, y]=\left(I_{in} \star h\right)[x, y]=\sum_{i \in \mathbb{Z}} \sum_{j \in \mathbb{Z}} I_{in}[x, y] h[i-x, j-y]
\end{equation}
where $\star $ means convolution, $I_{in}$, $I_{out}$ and $h$ represent the input image, output image, and learn-able convolutional kernel, respectively, $x,y\in \mathbb{Z}$ are pixel coordinates. Note that there are some slight differences from convolution in math to recent deep learning libraries \cite{abadi2016tensorflow, paszke2017automatic}, where channels, stride and padding are employed and no flip is required for convolutional kernels since they are randomly initialized. 

Batch normalization fixes the means and variances of each layer's inputs, it facilitates network convergence and alleviate over-fitting problem. Assume an input mini-batch with batch size $m$, for each dimension of BN layer’s input, the output is expressed as
\begin{equation}
\begin{split}
y & =\gamma \frac{x-\mu}{\sqrt{\sigma^{2}+\epsilon}} + \beta \\
\mu & =\frac{1}{m} \sum x \\
\sigma^{2} & =\frac{1}{m} \sum(x-\mu)^{2}
\end{split}
\end{equation}
where $x$ and $y$ are one dimension of BN layer’s input and output (both are $m$ dimensional vectors), $\mu $ and $\sigma ^2$ are mean and variance of input data, $\gamma \in \mathbb{R}$ and $\beta \in \mathbb{R}$ are learn-able parameters, a small constant $\epsilon $ is added for numerical stability.

The activation layer enables non-linearity of networks, which is attached after each layer in the network. We use Parametric ReLU (PReLU) \cite{he2015delving} expressed as
\begin{equation}
PReLU(x)=\left\{\begin{array}{ll}{x,} & {\text { if } x \geq 0} \\ {p x,} & {\text { otherwise }}\end{array}\right.
\end{equation}
where $p$ is a learn-able parameter, $x$ is the input.

Max pooling operation down sample an image while preserving its dominating features, expressed as
\begin{equation}
I_{out}[x, y]=\max _{i \in\{2 x-1,2 x\} \atop j \in\{2 y-1,2 y\}}\left(I_{in}[i, j]\right)
\end{equation}
where $I_{in}$ and $I_{out}$ represent the input image and output image, $x,y\in \mathbb{Z}$ are pixel coordinates.

Fully connected (FC) layers reduce or increase the dimension of features, using simple matrix product expressed as
\begin{equation}
V_{n}^{out}=W_{n \times m} \cdot V_{m}^{in}+b
\end{equation}
where $\cdot $ means matrix product, $V^{in}\in \mathbb{R}^m$ and $V^{out}\in \mathbb{R}^n$ are input and output vectors, $W \in \mathbb{R}^{n \times m}$ and $b \in \mathbb{R}$ are learn-able parameters. The first layer of FC requires huge number of learn-able parameters (about 134M) causing severe over-fitting problem, which is alleviated by dropout \cite{srivastava2014dropout}. Dropout disconnects the connections of neurons randomly with a fixed probability $p$. In this paper, dropout layers are attached after each FC layers with $p=0.5$ if the FC layer has more than 1M parameters.

The CNN is trained by minimizing the loss function below. We consider $B=\left(t_{x}, t_{y}, r_{z}, d_{x}, d_{y}\right)$ and $B=\left(t_{z}, d_{z}\right)$ in the first-step CNN and second-step CNN respectively.
\begin{equation}
E=\left\|\frac{1}{B_{gt}} \otimes\left(B_{pred}-B_{gt}\right)\right\|_{2}
\end{equation}
where $B_{gt}$ and $B_{pred}$ stand for ground truth bounding box and predicted bounding box respectively, the sign $\otimes $ is Hadamard product for vectors (i.e., element-wise product). The use of $\frac{1}{B_{gt}}$ forces the network to optimize smaller values in $B_{pred}$. Training data and details are described in Section 3.

\subsubsection{RoI in 2D: A foreground-background segmentation}
\label{sec:2_2_2}
\begin{figure}[htbp]	
	\centering	
	\includegraphics[width=\textwidth]{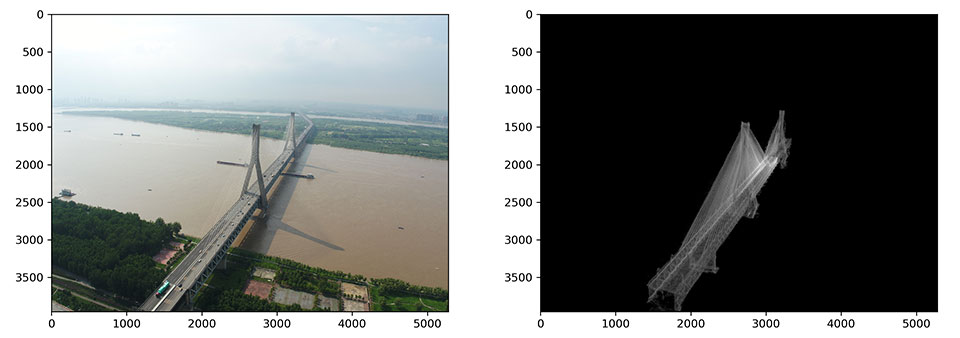}
	\caption{The original RGB image (left) and the obtained foreground-background segmentation (right) from 3D point cloud, 3D OBB and camera parameters.}
	\label{fig:segmentation}  
\end{figure}

RoI in 2D indicates a foreground-background segmentation mask (a gray scale image), it is obtained from 3D point cloud, 3D OBB and camera parameters, see Figure \ref{fig:segmentation}. 3D point cloud is first filtered by 3D OBB and then projected into 2D using camera parameters. The projection is expressed as
\begin{equation}
p^{T}=S_{\lambda} K[R | t] P^{T}
\end{equation}
where $P$ denotes filtered 3D point cloud, $S_{\lambda}$ is camera lens radial distortion parameters, $K$ and $[R | t]$ are intrinsic and extrinsic parameters, $p$ is the projection of $P$ on image coordinate with homogeneous representation. Note that $p$ is not ready to use, to get a foreground-background segmentation image, all 2D points in $p$ should be accumulated to 2D image grids. A general method is distance-based accumulation, when accumulating a 2D point with value 1, its four neighboring pixels are updated as $x_{i} :=x_{i}+d_{i} / \sum_{i=1}^{4} d_{i}$, where $x_i$ is the intensity value of a neighboring pixel, $d_i$ is the distance from 2D point to the neighboring pixel.

\subsection{Learning structural graph layouts and 3D shapes}
\label{sec:2_3}
\begin{figure}[htbp]
	\centering	
	\includegraphics[width=\textwidth]{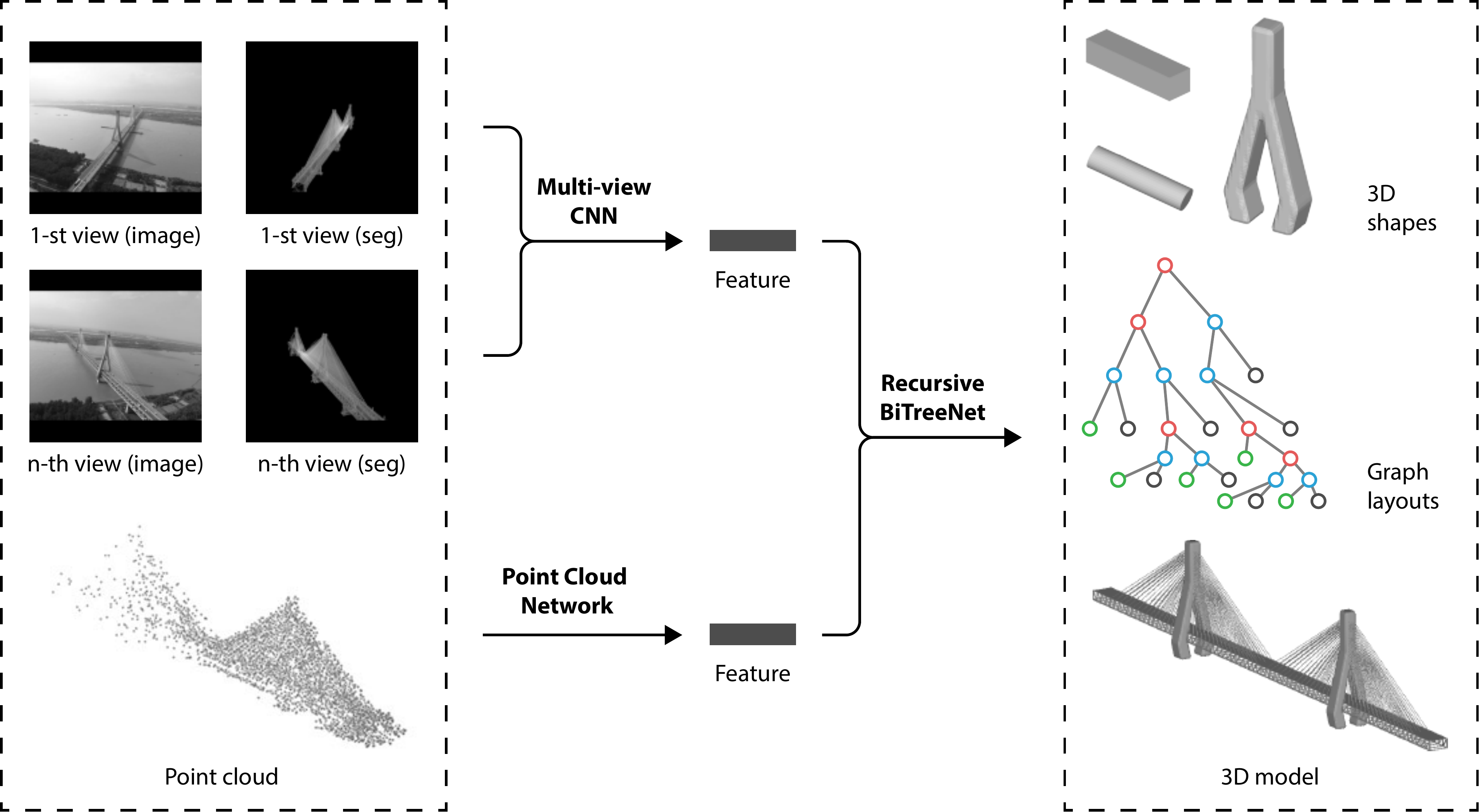}
	\caption{The proposed learning framework. Multi-view CNN and Point cloud network learn features from images and point clouds, Recursive BiTreeNet decodes features into the final 3D model.}
	\label{fig:full}  
\end{figure}

\begin{figure}[htbp]
	\centering	
	\includegraphics[width=\textwidth]{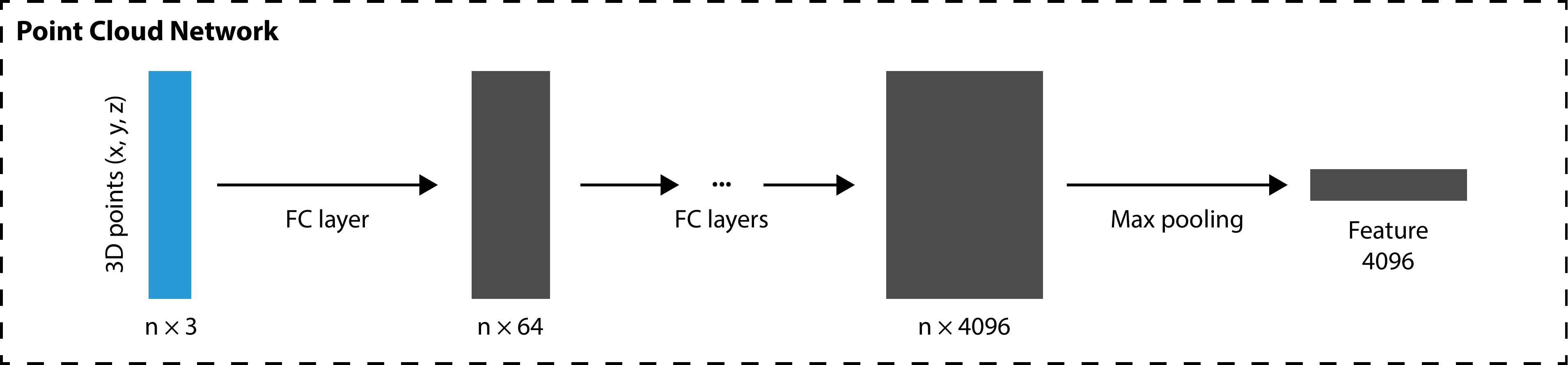}
	\caption{Point cloud network. Terminology: FC layer, fully connected layer; Max pooling, channel-wise (column-wise) max pooling.}
	\label{fig:pointnet}  
\end{figure}

This step is to learn structural graph layouts and 3D shapes from 3D point cloud and images. We design a learning framework, see Figure \ref{fig:full}. A multi-view convolutional neural network (Multi-view CNN, Figure \ref{fig:mvcnn}) combined with a point cloud network (Figure \ref{fig:pointnet}) is designed to encode multi-view images and point cloud into a latent feature, which is then decoded into structural graph layouts (i.e., a hierarchical structural parsing binary tree) by designing a recursive binary tree network (Recursive BiTreeNet, Figure \ref{fig:rvnn}), 3D shapes are decoded from shape nodes in the binary parsing tree (Figure \ref{fig:shape_decoder}). We formulate the learning framework below.

\subsubsection{Multi-view convolutional neural network (Multi-view CNN)}
\label{sec:2_3_1}
\begin{figure}[htbp]
	\centering	
	\includegraphics[width=\textwidth]{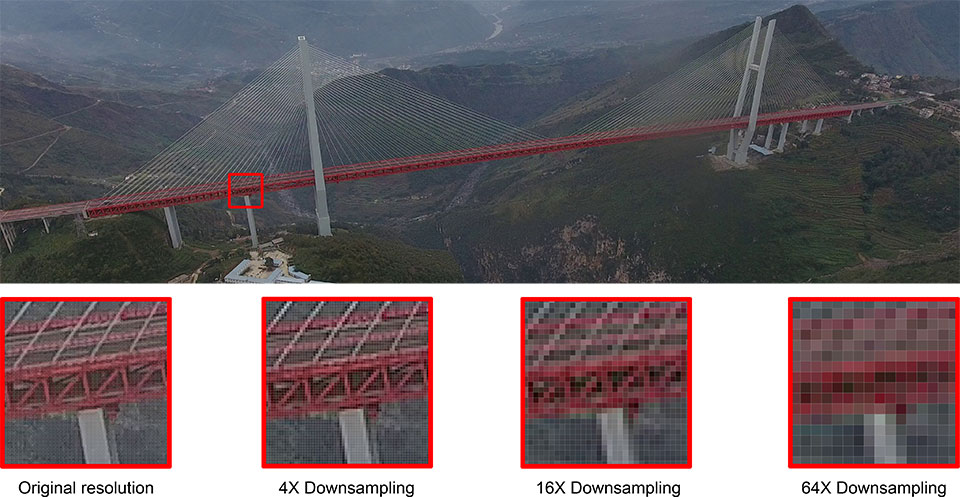}
	\caption{Choosing a resolution. 16 times down-sampling is chosen as a balance between memory cost and performance.}
	\label{fig:resolution}  
\end{figure}

\begin{figure}[htbp]
	\centering	
	\includegraphics[width=\textwidth]{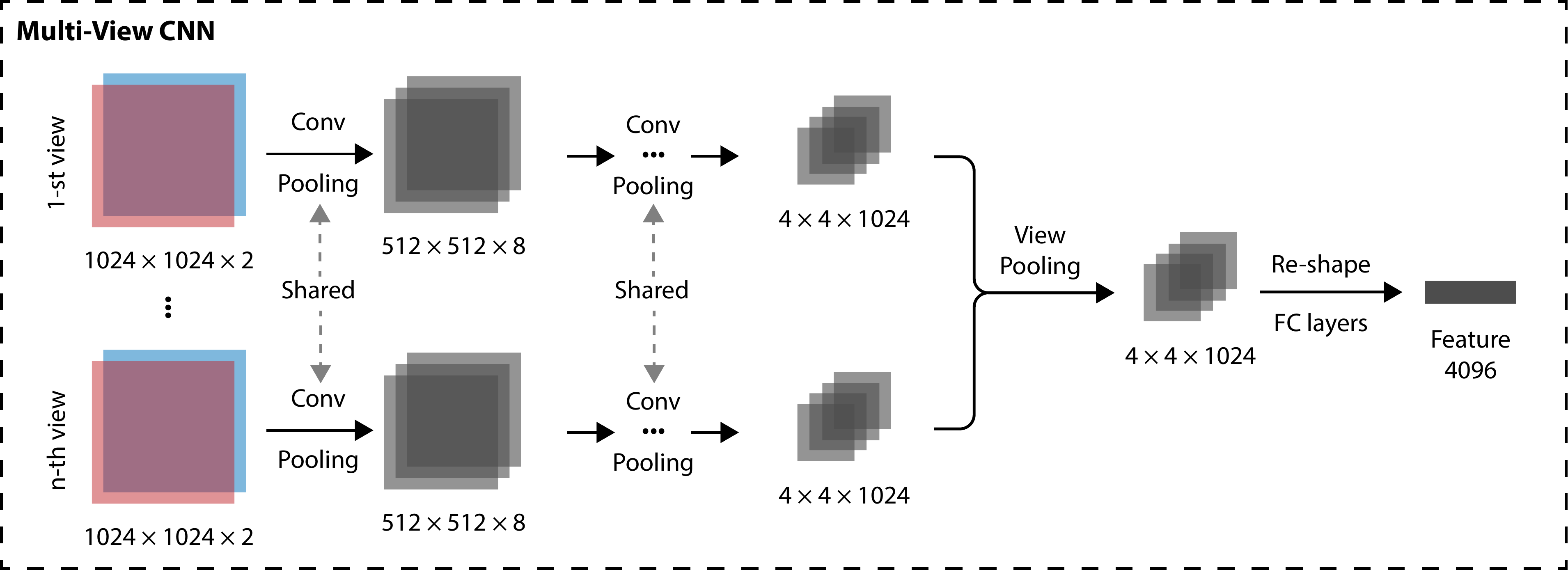}
	\caption{Multi-view Convolutional Neural Network (Multi-view CNN). Terminology: Conv, convolutional layer; Pooling, max pooling layer; Re-shape, convert a tensor to a vector; FC layers, fully connected layers, indicating a three-layer fully connected network in this figure.}
	\label{fig:mvcnn}  
\end{figure}

CNNs learn features from images. A multi-view CNN in this section (Figure \ref{fig:mvcnn}) learns a feature $h\in \mathbb{R}^{4096}$ from $n$-view images $X=\left\{I_{i} \in \mathbb{N}^{1024 \times 1024 \times 2} | i=1, \ldots, n\right\}$, the Multi-view CNN can be seen as a mapping function $f:X\to h$. A single view $I_i$ has two channels, a gray scale image and its corresponding foreground-background segmentation. A 16 times down sampled image resolution (about $1000\times 1000$) was chosen to balance algorithm performance and memory overhead, see Figure \ref{fig:resolution}. Such high-resolution images with multiple views require huge learning parameters and memory, to alleviate this, we use multi-view CNN, where convolutional kernels among different views are shared, compared with general framework like VGG16 \cite{simonyan2014very} (553M), ResNet101 \cite{he2016deep} (178M) and DenseNet121 \cite{huang2017densely} (31M), the proposed multi-view CNN (6M) reduced learning parameters and memory cost significantly, thus works on a consumer grade GPU with 11GB memory and with batch size 1 up to 6 views.

Figure \ref{fig:mvcnn} illustrates the Multi-view CNN, each view branch contains convolutional layers, activation layers and max pooling layers. Batch normalization layers are not considered in this section since all inputs are with batch size 1. Learn-able parameters are shared among different view branches. The view pooling layer compares features from different views and selects maximum features, expressed as $I_{out}[x, y, c]=\max _{m \in\{1, \ldots, n\}}\left(I_{in}[x, y, c, m]\right)$, where $I_{in}$ and $I_{out}$ represent the input and output feature, $x,y\in \mathbb{Z}$ denotes pixel coordinate, $c\in \mathbb{Z}$ is channel index, $m\in \{1,...,n\}$ is view index for $n$ input views. The output feature is then re-shaped to a vector and fed into three fully connected layers. The final feature is a 4096-dimensional vector. Table \ref{tab:mvcnn} lists the detailed properties of layers and operators.

\begin{table}[htbp]
	\centering
	\caption{Detailed properties of layers and operators.}
	\label{tab:mvcnn}
	\setlength\heavyrulewidth{0.35ex}
	\begin{tabular}{lllllll}
		\toprule
		Layers & Feature size                & Operators     & Kernel size              & No.  & Stride & Padding \\
		\midrule
		0      & $1024 \times 1024 \times 2$ & Input         & -                        & -    & -      & -       \\
		1      & $1024 \times 1024 \times 8$ & Conv          & $3 \times 3 \times 2$    & 8    & 1      & 1       \\
		2      & $1024 \times 1024 \times 8$ & PReLU         & -                        & -    & -      & -       \\
		3      & $512 \times 512 \times 8$   & MP            & $2 \times 2 \times 8$    & -    & 2      & -       \\
		4      & $512 \times 512 \times 16$  & Conv          & $3 \times 3 \times 8$    & 16   & 1      & 1       \\
		5      & $512 \times 512 \times 16$  & PReLU         & -                        & -    & -      & -       \\
		6      & $256 \times 256 \times 16$  & MP            & $2 \times 2 \times 16$   & -    & 2      & -       \\
		7      & $256 \times 256 \times 32$  & Conv          & $3 \times 3 \times 16$   & 32   & 1      & 1       \\
		8      & $256 \times 256 \times 32$  & PReLU         & -                        & -    & -      & -       \\
		9      & $128 \times 128 \times 32$  & MP            & $2 \times 2 \times 32$   & -    & 2      & -       \\
		10     & $128 \times 128 \times 64$  & Conv          & $3 \times 3 \times 32$   & 64   & 1      & 1       \\
		11     & $128 \times 128 \times 64$  & PReLU         & -                        & -    & -      & -       \\
		12     & $64 \times 64 \times 64$    & MP            & $2 \times 2 \times 64$   & -    & 2      & -       \\
		13     & $64 \times 64 \times 128$   & Conv          & $3 \times 3 \times 64$   & 128  & 1      & 1       \\
		14     & $64 \times 64 \times 128$   & PReLU         & -                        & -    & -      & -       \\
		15     & $32 \times 32 \times 128$   & MP            & $2 \times 2 \times 128$  & -    & 2      & -       \\
		16     & $32 \times 32 \times 256$   & Conv          & $3 \times 3 \times 128$  & 256  & 1      & 1       \\
		17     & $32 \times 32 \times 256$   & PReLU         & -                        & -    & -      & -       \\
		18     & $16 \times 16 \times 256$   & MP            & $2 \times 2 \times 256$  & -    & 2      & -       \\
		19     & $16 \times 16 \times 512$   & Conv          & $3 \times 3 \times 256$  & 512  & 1      & 1       \\
		20     & $16 \times 16 \times 512$   & PReLU         & -                        & -    & -      & -       \\
		21     & $8 \times 8 \times 512$     & MP            & $2 \times 2 \times 512$  & -    & 2      & -       \\
		22     & $8 \times 8 \times 1024$    & Conv          & $3 \times 3 \times 512$  & 1024 & 1      & 1       \\
		23     & $8 \times 8 \times 1024$    & PReLU         & -                        & -    & -      & -       \\
		24     & $4 \times 4 \times 1024$    & MP            & $2 \times 2 \times 1024$ & -    & 2      & -       \\
		25     & $4 \times 4 \times 1024$    & View pooling  & -                        & -    & -      & -       \\
		26     & $16384$                     & Re-shape      & -                        & -    & -      & -       \\
		27     & $10321$                     & FC            & $16384 \times 10321$     & -    & -      & -       \\
		28     & $10321$                     & Dropout (0.5) & -                        & -    & -      & -       \\
		29     & $10321$                     & PReLU         & -                        & -    & -      & -       \\
		30     & $6502$                      & FC            & $10321 \times 6502$      & -    & -      & -       \\
		31     & $6502$                      & Dropout (0.5) & -                        & -    & -      & -       \\
		32     & $6502$                      & PReLU         & -                        & -    & -      & -       \\
		33     & $4096$                      & FC            & $6502 \times 4096$       & -    & -      & -       \\
		34     & $4096$                      & Dropout (0.5) & -                        & -    & -      & -       \\
		35     & $4096$                      & Output        & -                        & -    & -      & -       \\
		\bottomrule
	\end{tabular}
\end{table}

\subsubsection{Point cloud network}
\label{sec:2_3_2}
Point cloud network learns features from a point cloud (a point set). A point cloud network in this section learns a feature $h\in \mathbb{R}^{4096}$ from a 3D point set $X=\{P_i \in  \mathbb{R}^3 |i=1,...,n\}$, the point cloud network can be seen as a mapping function $f:X\to h$. A point cloud is composed by a set of points with following properties: (a) disorder; (b) unfixed number; (c) arbitrary rotation. Based on (a) and (b), symmetry functions (e.g., max function where $max(x,y)=max(y,x)$) are introduced \cite{qi2017pointnet}. Unlike recent popular methods \cite{qi2017pointnet, qi2017pointnet++, li2018pointcnn}, the arbitrary rotation was already solved in Section 2.2. To solve (a) and (b), given an input point cloud $\{P_i\}_{i=1}^n$ composed of $n$ 3D points, a series of fully connected layers followed by a max pooling layer are employed in this section. Table \ref{tab:pointnet} lists the detailed properties of layers and operators. We will describe these layers in the following.

\begin{table}[htbp]
	\centering
	\caption{Detailed properties of layers and operators.}
	\label{tab:pointnet}
	\setlength\heavyrulewidth{0.35ex}
	\begin{tabular}{llll}
		\toprule
		Layers & Feature size    & Operators     & Kernel size        \\
		\midrule
		0      & $n \times 3$    & Input         & -                  \\
		1      & $n \times 64$   & FC            & $3 \times 64$      \\
		2      & $n \times 64$   & PReLU         & -                  \\
		3      & $n \times 256$  & FC            & $64 \times 256$    \\
		4      & $n \times 256$  & PReLU         & -                  \\
		5      & $n \times 1024$ & FC            & $256 \times 1024$  \\
		6      & $n \times 1024$ & PReLU         & -                  \\
		7      & $n \times 4096$ & FC            & $1024 \times 4096$ \\
		8      & $n \times 4096$ & Dropout (0.5) & -                  \\
		9      & $n \times 4096$ & PReLU         & -                  \\
		10     & $4096$          & MP            & $4096$             \\
		11     & $4096$          & Output        & -                  \\
		\bottomrule
	\end{tabular}
\end{table}

Fully connected layer. The unordered input requires shared weights among different points, meanwhile allowing arbitrary number of points. Hence a fully connected layer meets these requirements, which indicates matrix product expressed as
\begin{equation}
P^{out}_{n \times t} = P^{in}_{n \times s} \cdot W_{s \times t} + b
\end{equation}
where $\cdot$ means matrix product, $P^{in} \in \mathbb{R}^{n \times s}$ and $P^{out} \in \mathbb{R}^{n \times t}$ denote $n$ input and output points in Euclidean $s$-space and $t$-space, respectively, forming two matrixes, $W \in \mathbb{R}^{s \times t}$ and $b \in \mathbb{R}$ are learn-able parameters.

Channel-wise (column-wise) max pooling layer. The channel-wise max pooling in this section indicates the symmetry function in \cite{qi2017pointnet}, where a max value is selected in each channel (column) for $n$ points, expressed as
\begin{equation}
h[x] = \max \limits_{i \in \{1,...,n\}} P^{in}_{n \times s}[i, x]
\end{equation}
where $P^{in} \in \mathbb{R}^{n \times s}$ denotes $n$ input points in Euclidean $s$-space, $h \in \mathbb{R}^{s}$ denotes the learned feature, $x \in \mathbb{Z}$ is the channel index..

\subsubsection{Recursive binary tree network (Recursive BiTreeNet)}
\label{sec:2_3_3}
A binary parsing tree is a rooted tree, where every node has at most two children, it is defined as a graph $\mathcal{T} = \{V, E_1 \cup E_2\}$ and $E_1 \cap E_2$ is empty. The notion of parsing tree comes from the field of Linguistics and becomes frequently used in learning-based natural language processing \cite{socher2011parsing, socher2013recursive} and computer graphics research \cite{choi2013understanding, li2017grass}. For example, a sentence \textit{"The cat sleeps on the carpet."} is parsed into \textit{"(ROOT (S (NP (DT The) (NN cat)) (VP (VBD sleeps) (PP (IN on) (NP (DT the) (NN carpet)))) (. .)))"}, where $\{\textit{"ROOT", "NP", ..., "NN"}\}$ are node types in the binary parsing tree.

\begin{figure}[htbp]
	\centering	
	\includegraphics[width=0.6\textwidth]{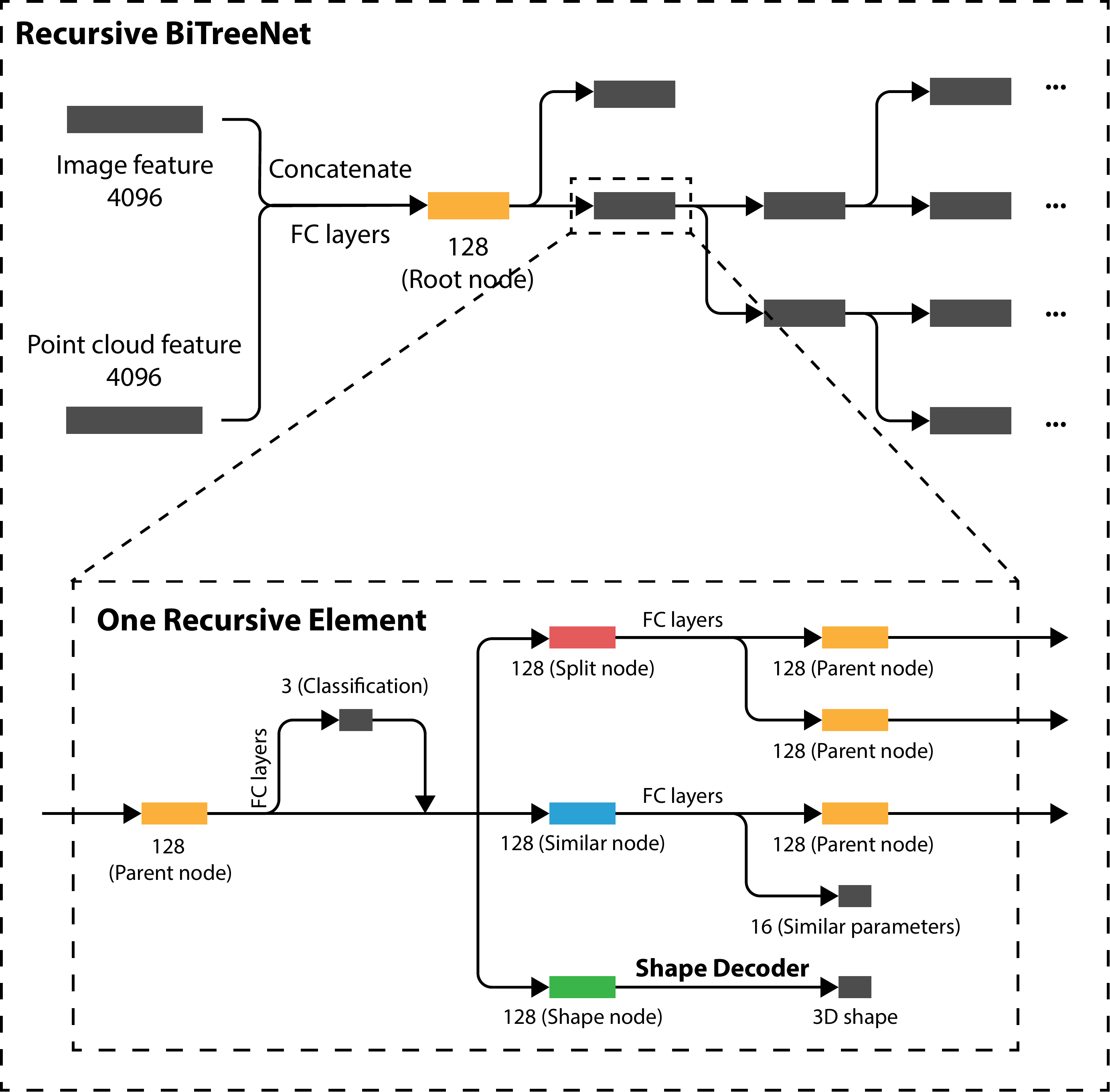}
	\caption{Recursive Binary Tree Network (Recursive BiTreeNet). Terminology: Concatenate, concatenating two vectors; FC layers, fully connected layers, all FC layers in this figure indicate a two-layer fully connected network.}
	\label{fig:rvnn}
\end{figure}

The proposed recursive binary tree network (see Figure \ref{fig:rvnn}) learns a binary tree from image features and point cloud features. The binary tree describes the structural graph layouts, where the shape nodes can be seen as 3D shape codes and are decode into 3D shapes by shape decoder, and similar nodes describe the distribution of 3D shapes.

In Figure \ref{fig:rvnn}, the Recursive BiTreeNet first fuse image feature and point cloud feature by two FC layers, the output is a 128-D root node feature. The root node is regarded as a parent node and is decoded to a binary tree by the BiTreeNet recursively. One recursive element is shown in Figure \ref{fig:rvnn}, a parent node is decoded to one or two child nodes, and the recursive element is reused in the next hierarchy. Given a root node $X\in \mathbb{R}^{128}$, the recursion processing is written in Algorithm \ref{algorithm}. We will explain Algorithm \ref{algorithm} below in detail.

\begin{algorithm}[H]
	\setstretch{2.5}
	\SetAlgoLined
	\DontPrintSemicolon
	
	\SetKwFunction{FMain}{Recursion}
	\SetKwProg{Fn}{Function}{:}{}
	\Fn{\FMain{$X^i \in \mathbb{R}^{128}$, $X_{cls}^{i} \in \{0,1,2,None\}$}}{
		
		\uIf{$X_{cls}^{i}=None$}{
			$X_{cls}^{i}$ $\leftarrow$ NodeClassifier($X^i$)
		}  
	
		\uElseIf{$X_{cls}^{i}=0$}{
			($X_{left}^{i+1} \in \mathbb{R}^{128}$, $X_{right}^{i+1}\in \mathbb{R}^{128}$) $\leftarrow$ SplitNode($X^i$)
			\\
			\textbf{return} Recursion($X_{left}^{i+1}$, $X_{left\_cls}^{i+1}$) and Recursion($X_{right}^{i+1}$, $X_{right\_cls}^{i+1}$)
		}
	
		\uElseIf{$X_{cls}^{i}=1$}{
			($X_{left}^{i+1} \in \mathbb{R}^{128}$, $X_{right}^{i+1}\in \mathbb{R}^{16}$) $\leftarrow$ SimilarNode($X^i$)
			\\
			\textbf{return} Recursion($X_{left}^{i+1}$, $X_{left\_cls}^{i+1}$) and $X_{right}^{i+1}$
		}
	
		\Else{
			$X^{i+1}$ $\leftarrow$ ShapeNode($X^i$)
			\\
			\textbf{return} $X^{i+1}$
		}  
					 
	}
	\textbf{End Function}
	\caption{One recursive element function.}
	\label{algorithm}
\end{algorithm}

Algorithm \ref{algorithm} shows the recursion function. In one recursive element, in training stage, the node class is known ($X_{cls}^{i}\in \{0,1,2\}$), but not in test stage ($X_{cls}^{i}=None$). The node classifier first classifies the parent node into three node types, for training stage, we use the known node class and calculate node classification loss; for test stage, we use the node class given by node classifier $X_{cls}^{i} \leftarrow \text{NodeClassifier}(X^i)$. The node classifier is a two-layer fully connected network, the output is a 3-D one-hot vector indicating probabilities of 3 node types, e.g., $X_{cls}^{i}=(0,1,0)$ means the second type and is equivalent to $X_{cls}^{i}=1$, we use the latter in written form to simplify. The three node types are split node, similar node and shape node, Figure \ref{fig:binary_tree} shows functions of these nodes. 

$X_{cls}^{i}=0$ means the current node is a split node, and it can split into 2 nodes, that is, the current shape can split into 2 adjacent shapes, realized by a two-layer FC network $f:\mathbb{R}^{128}\to \mathbb{R}^{128+128}$. The two output child nodes can be seen as two parent nodes for the next hierarchy, they are then fed into the next recursion function.

$X_{cls}^{i}=1$ means the current node is a similar node, and it can split into 2 nodes, which means the current shape can split into one shape and its "copies", obtained by a two-layer FC network $f:\mathbb{R}^{128}\to \mathbb{R}^{128+16}$. The two output child nodes can be seen as one parent node for the next hierarchy, and one similar node with 16 similar parameters. We use up to 16 parameters to represent similarity, the first 3 parameters are a one-hot vector indicating the similar type, including 1-D translation (rigid translation), 2-D translation (non-rigid translation) and reflection. For 1-D translation (e.g., truss element), 1 parameter is for copy number, 1 parameter is for distance, 3 parameters are for translation directions specified by a 3-D vector, e.g., $(1,0,0)$ means the $x$ direction. For 2-D translation, e.g., cables are similar but vary in lengths, 1 parameter is for copy number, 2 parameters are for 2 endpoints’ translation distances, 6 parameters are for 2 endpoints’ translation directions specified by 2 3-D vectors. For reflection, 1 parameter is for reflection distance, 3 parameters are for reflection directions specified by a 3-D vector. The rest parameters are set to 0 in labels. Here we use a small trick to predict a reasonable 3D model, notice that translation distance of cable and truss are correlative, but there is no edge between similar nodes of cables and trusses. Actually, this correlation hasn’t been modeled in binary tree, we simply merged close translation distance using their average, e.g., two translation distances are 0.121 and 0.117, we use 0.119 for both.

$X_{cls}^{i}=2$ indicates that the current node is a shape node, which means the current shape is inseparable. The shape node is decoded to a 3D shape by Shape decoder in Section \ref{sec:2_3_4}. 

One may notice that these operations require dynamic computation graph, which means the computation graph is variable during each iteration. This is friendly supported in \cite{paszke2017automatic} but not in \cite{abadi2016tensorflow}. Table \ref{tab:rvnn} lists the detailed properties of layers and operators.

\begin{table}[htbp]
	\centering
	\caption{Detailed properties of layers and operators.}
	\label{tab:rvnn}
	\setlength\heavyrulewidth{0.35ex}
	\begin{tabular}{llll}
		\toprule
		Layers         & Feature size  & Operators     & Kernel size        \\
		\midrule
		\multirow{7}{*}{Feature fusion} 
		& 4096+4096    & Input         & -                  \\
		& 8192         & Concatenation & -                  \\
		& 1024         & FC            & $8192 \times 1024$ \\
		& 1024         & Dropout       & -                  \\
		& 1024         & PReLU         & -                  \\
		& 128          & FC            & $1024 \times 128$  \\
		& 128          & PReLU         & -                  \\
		\midrule
		\multirow{5}{*}{NodeClassifier} 
		& 128          & Input         & -                  \\
		& 20           & FC            & $128 \times 20$    \\
		& 20           & PReLU         & $20$               \\
		& 3            & FC            & $20 \times 3$      \\
		& 3            & Softmax       & $3$                \\
		\midrule
		\multirow{5}{*}{SplitNode}      
		& 128          & Input         & -                  \\
		& 181          & FC            & $128 \times 181$   \\
		& 181          & PReLU         & -                  \\
		& 256          & FC            & $181 \times 256$   \\
		& 128+128      & Output        & -                  \\
		\midrule
		\multirow{5}{*}{SimilarNode}    
		& 128          & Input         & -                  \\
		& 136          & FC            & $128 \times 136$   \\
		& 136          & PReLU         & -                  \\
		& 144          & FC            & $136 \times 144$   \\
		& 128+16       & Output        & -                  \\
		\midrule
		\multirow{2}{*}{ShapeNode}      
		& 128          & Input         & -                  \\
		& A 3D shape   & Output        & -                  \\
		\bottomrule
	\end{tabular}
\end{table}

\subsubsection{Shape decoder network}
\label{sec:2_3_4}
\begin{figure}[htbp]
	\centering	
	\includegraphics[width=\textwidth]{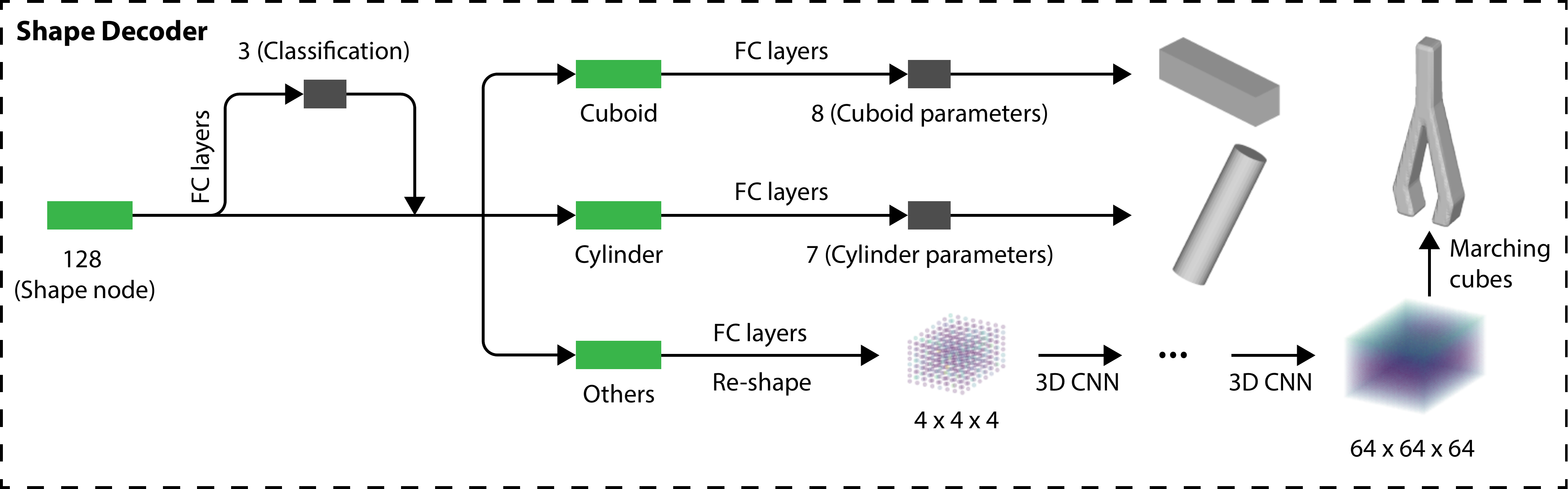}
	\caption{Shape decoder. Terminology: FC layers, fully connected layer; Re-shape, convert a vector to a tensor; 3D CNN, 3D convolutional layer; Marching cubes, differentiable marching cubes (MC) algorithm \cite{liao2018deep}.}
	\label{fig:shape_decoder}  
\end{figure}

Shape decoder network (Figure \ref{fig:shape_decoder} is to decode a 128-D shape node vector into a 3D shape. The shape decoder first classifies the input shape node into 3 types, including cuboid, cylinder and irregular shapes like bridge tower. Similar to the node classifier mentioned above in Section \ref{sec:2_3_3}, for training stage, we use the known shape type and calculate node classification loss; for test stage, we use the shape type given by node classifier.

A cuboid requires 8 parameters, $2 \times 3=6$ for 2 endpoints and 2 for section width and height, obtained from a two-layer FC network $f:\mathbb{R}^{128}\to \mathbb{R}^{8}$. A cylinder requires 7 parameters, $2 \times 3=6$ for 2 endpoints and 1 for section radius. Table \ref{tab:shape_decoder} lists detailed properties of layers and operators except for 3D CNNs.

\begin{table}[htbp]
	\centering
	\caption{Detailed properties of layers and operators.}
	\label{tab:shape_decoder}
	\setlength\heavyrulewidth{0.35ex}
	\begin{tabular}{llll}
		\toprule
		Layers                          & Feature size             & Operators & Kernel size     \\
		\midrule
		\multirow{5}{*}{NodeClassifier} 
		& 128                      & Input     & -               \\
		& 20                       & FC        & $128 \times 20$ \\
		& 20                       & PReLU     & $20$            \\
		& 3                        & FC        & $20 \times 3$   \\
		& 3                        & Softmax   & $3$             \\
		\midrule
		\multirow{5}{*}{Cuboid}         
		& 128                      & Input     & -               \\
		& 32                       & FC        & $128 \times 32$ \\
		& 32                       & PReLU     & -               \\
		& 8                        & FC        & $32 \times 8$   \\
		& 8                        & Output    & -               \\
		\midrule
		\multirow{5}{*}{Cylinder}       
		& 128                      & Input     & -               \\
		& 30                       & FC        & $128 \times 30$ \\
		& 30                       & PReLU     & -               \\
		& 7                        & FC        & $30 \times 7$   \\
		& 7                        & Output    & -               \\
		\midrule
		\multirow{2}{*}{Irregular}      
		& 128                      & Input     & -               \\
		& $64 \times 64 \times 64$ & Output    & -               \\
		\bottomrule            
	\end{tabular}
\end{table}

For an irregular shape, 3D CNNs are employed to decode the shape node to a 3D occupancy grid $O\in \mathbb[0,1]^{64\times 64\times 64}$ and a vertex displacement grid $X\in \mathbb[0,1]^{64\times 64\times 64\times 3}$, then differentiable (MC) converts these two grids to a triangular mesh. A 128-D shape node is first mapped to a $4\times 4\times 4$ grid with 512 channels by a FC layer, the voxel grid is then up-sampled by a series of fractional-stride convolutional layers. Fractional-stride convolution first adds spacing between existing pixels (or voxels) within a feature map to increase spatial resolution, the spacing are then filled with zeros, the rest is to do regular convolution operation. Fractional-stride convolution is also known as transposed convolution or dilated convolution, the spacing between pixels is called dilation. Table \ref{tab:3dcnn} lists the detailed properties of layers and operators in 3D fractional-stride CNN.

\begin{table}[htbp]
	\centering
	\caption{Detailed properties of layers and operators in 3D fractional-stride CNN.}
	\label{tab:3dcnn}
	\setlength\heavyrulewidth{0.35ex}
	\begin{tabular}{llllllll}
		\toprule 
		Layers & Feature size                          & Operators         & Kernel size                      & No. & Dilation & Stride & Padding \\
		\midrule
		0      & $128$                                 & Input             & -                                & -   & -        & -      & -       \\
		1      & $64 \times 512$                       & FC                & $128 \times 64$                  & 512 & -        & -      & -       \\
		2      & $64 \times 512$                       & PReLU             & -                                & -   & -        & -      & -       \\
		3      & $4 \times 4 \times 4 \times 512$      & Re-shape          & -                                & -   & -        & -      & -       \\
		4      & $8 \times 8 \times 8 \times 256$      & Transposed Conv   & $4 \times 4 \times 4 \times 512$ & 256 & 1        & 1      & 2       \\
		5      & $8 \times 8 \times 8 \times 256$      & PReLU             & -                                & -   & -        & -      & -       \\
		6      & $16 \times 16 \times 16 \times 128$   & Transposed Conv   & $4 \times 4 \times 4 \times 256$ & 128 & 1        & 1      & 2       \\
		7      & $16 \times 16 \times 16 \times 128$   & PReLU             & -                                & -   & -        & -      & -       \\
		8      & $32 \times 32 \times 32 \times 64$    & Transposed Conv   & $4 \times 4 \times 4 \times 128$ & 64  & 1        & 1      & 2       \\
		9      & $32 \times 32 \times 32 \times 64$    & PReLU             & -                                & -   & -        & -      & -       \\
		10     & $64 \times 64 \times 64 \times (1+3)$ & Transposed Conv   & $4 \times 4 \times 4 \times 64$  & 4   & 1        & 1      & 2       \\
		11     & $64 \times 64 \times 64 \times (1+3)$ & Softmax           & -                                & -   & -        & -      & -       \\
		12     & Mesh                                  & Differentiable MC & -                                & -   & -        & -      & -       \\
		\bottomrule
	\end{tabular}
\end{table}

To enable end-to-end training, the unsigned distance field is converted to a triangular mesh through differentiable marching cubes \cite{liao2018deep}. The final 3D model is obtained by "copying" the decoded 3D shapes with similar parameters.

\subsubsection{Loss functions}
\label{sec:2_3_5}
The structural graph layouts are modeled by minimizing the summation of cross entropy classification loss $L_{cls}$ for each node expressed as
\begin{equation}
L_{cls}=-\frac{1}{n} \sum_{i \in\{1, \ldots, n\}} \sum_{x \in X} p_{i}(x) \log q_{i}(x)
\end{equation}
where $p_{i}(x)\in \{0,1\}$ and $q_{i}(x)\in (0,1]$ denote ground truth probability and predicted probability for event $x$ in events set $X$ for the $i$-th node in total $n$ nodes. The event $x$ means, for example, if the current node is a first type node, then $p_{i}(x)$ is either 0 or 1, $q_{i}(x)$ is the first value in the one-hot 3-D vector. It should be note that the binary tree including node types is known in training procedure, hence the node classifier is trained in training and is useful only when testing.

3D shapes are modeled by minimizing the distance loss function $L_{shape}$ for all shapes (taken as meshes) decoded by shape decoder expressed as
\begin{equation}
L_{shape}=\frac{1}{n} \sum_{i \in\{1, \ldots, n\}} dist \left(M_{i}^{gt}, M_{i}^{pred}\right)
\end{equation}
where $M_{i}^{gt}$ and $M_{i}^{pred}$ are ground truth shape and predicted shape in the $i$-th shape node in total $n$ shape nodes, $dist$ is Chamfer distance for the case that two shapes have different number of vertices, expressed as
\begin{equation}
dist\left(M_{1}, M_{2}\right)=\frac{1}{m_{1}} \sum_{v_{1} \in M_{1}} \min _{v_{2} \in M_{2}}\left\|v_{1}-v_{2}\right\|_{2}+\frac{1}{m_{2}} \sum_{v_{2} \in M_{2}} \min _{v_{1} \in M_{1}}\left\|v_{1}-v_{2}\right\|_{2}
\end{equation}
where $v_{1}$ and $v_{2}\in \mathbb{R}^{3}$ are vertices in shapes $M_{1}$ and $M_{2}$ respectively, $m_{1}$ and $m_{2}$ are vertices number in $M_{1}$ and $M_{2}$. For cuboids, 8 vertices are parameterized with 8 parameters mentioned above, for cylinders, we use cuboids with square section for calculation and back to cylinders for visualization. No smooth or long edge length regularization terms are considered since the mesh is converted from a regular voxel grid.

The final 3D model requires to minimize the similar nodes loss function $L_{sim}$
\begin{equation}
L_{sim}=\frac{1}{n} \sum_{i \in\{1, \ldots, n\}}\left\|S_{i}^{pred}-S_{i}^{gt}\right\|_{2}
\end{equation}
where $S_{i}^{gt}$ and $S_{i}^{pred}\in \mathbb{R}^{16}$ are the $i$-th ground truth similar parameters and the $i$-th predicted parameters in total $n$ similar nodes.

The total loss function $L$ is the summation of all mentioned losses expressed as
\begin{equation}
L=\lambda_{cls} L_{cls} + \lambda_{shape} L_{shape} + \lambda_{sim} L_{sim}
\end{equation}
where $\lambda_{cls}$, $\lambda_{shape}$ and $\lambda_{sim}$ are 3 weights to compose the total loss. The overall loss $L$ is to update the network in an end-to-end fashion.

\section{EXPERIMENTS}
\subsection{Training Details}
\begin{figure}[htbp]
	\centering	
	\includegraphics[width=\textwidth]{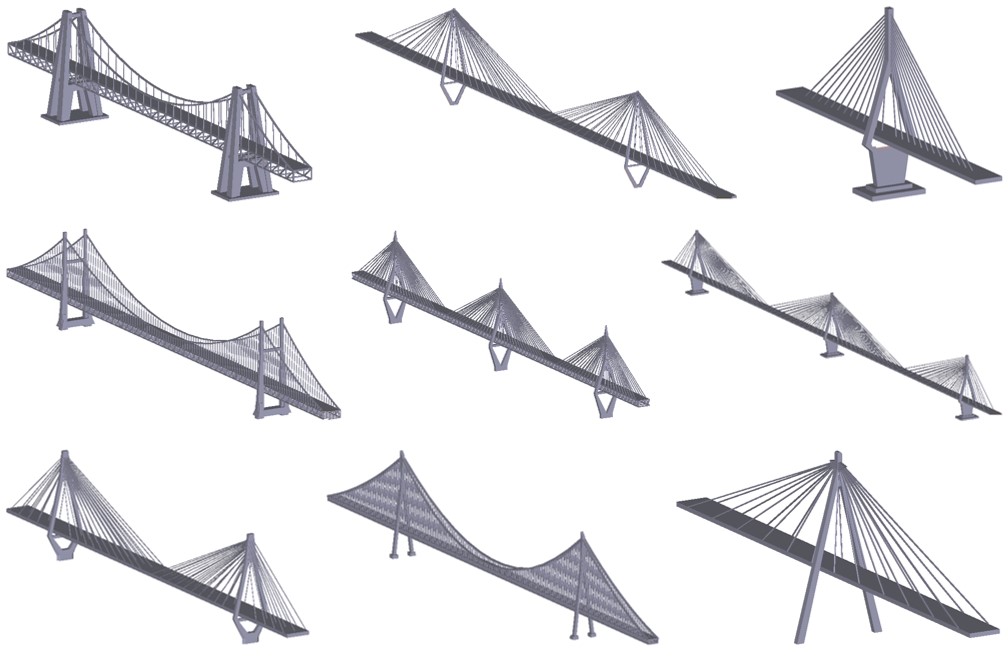}
	\caption{Some training samples.}
	\label{fig:training_sample}  
\end{figure}

We collected and generated about 2200 models of long span bridges, see Figure \ref{fig:training_sample}. Each training sample includes inputs and labels.

\subsubsection{Details for 3D OBB learning}
The point clouds are synthesized by densely sampled 3D bridge point cloud (about 1M points) with Gaussian noise and non-uniform sampling, and random terrain point cloud (about 6M points). The synthesized point clouds are first filtered with a sphere boundary mentioned in Section \ref{sec:2_2_1}. Bird views are generated by projecting point cloud within the sphere boundary to an image along $z$ (up) direction, front views are generated by projecting point cloud within the 2D OBB $B=(t_x,t_y,r_z,d_x,d_y)$ to an image along its local $y$ (front) direction. The labels are 3D OBBs $B=(t_x,t_y,r_z,d_x,d_y,t_z,d_z)$. Images are with size $1024\times 1024\times 1$ and are normalized to $(0,1)$, point cloud within the sphere boundary is translated and scaled to a unit sphere with center $(0,0,0)$.

Learning rate is set to $1\times 10^{-2}$ initially and $1\times 10^{-6}$ finally with exponential decline, training epoch is 200 with batch size 3, we use adaptive moment estimation (Adam) with $\beta_{1}=0.9$ and $\beta_{2}=0.999$ to optimize the network, no pre-training is implemented.

\subsubsection{Details for structural graph layouts and 3D shapes learning}
The inputs are composed of: (a) a sparsely sampled 3D point cloud (from 4K to 8K points) with Gaussian noise and non-uniform sampling; (b) 4 to 8 views, each view includes a rendered gray scale image from 3D model with random background, and a re-projected foreground-background segmentation from a densely sampled 3D point cloud (about 1M points) with Gaussian noise and non-uniform sampling. Images are with size $1024\times 1024\times 1$ and are normalized to $(0,1)$, the geometric centers of point clouds are translated to $(0,0,0)$, point clouds are scaled to 1 according to $z$ (up) direction.

The labels are composed of: (a) a binary tree represented with nested tuples (in Python language) by assembling one of the three node types on each node; (b) the 3D model with 3D shapes (mesh) on shape nodes and similar parameters on similar nodes.

Learning rate is set to $1\times 10^{-3}$ initially and $1\times 10^{-6}$ finally with exponential decline, training epoch is 500 with batch size 1, we use adaptive moment estimation (Adam) with $\beta_{1}=0.9$ and $\beta_{2}=0.999$ to optimize the network, no pre-training is implemented. In the loss function, we choose $\lambda_{cls}=1$, $\lambda_{shape}=20$ and $\lambda_{sim}=5$.

\subsection{Qualitative comparison}
In addition to these methods, we manually create a 3D model based on point cloud and images as a reference, which is regarded as the best w.r.t details but not in accuracy, however, it takes a dozen of hours for manual work. It's worth mentioning that when manually creating these models, we refer to three orthographic views of point cloud to obtain approximate size information, and refer to raw images to infer detailed part relations and 3D shapes, hence it is reasonable that how the learning framework works. 

\begin{figure}[htbp]
	\centering	
	\includegraphics[width=\textwidth]{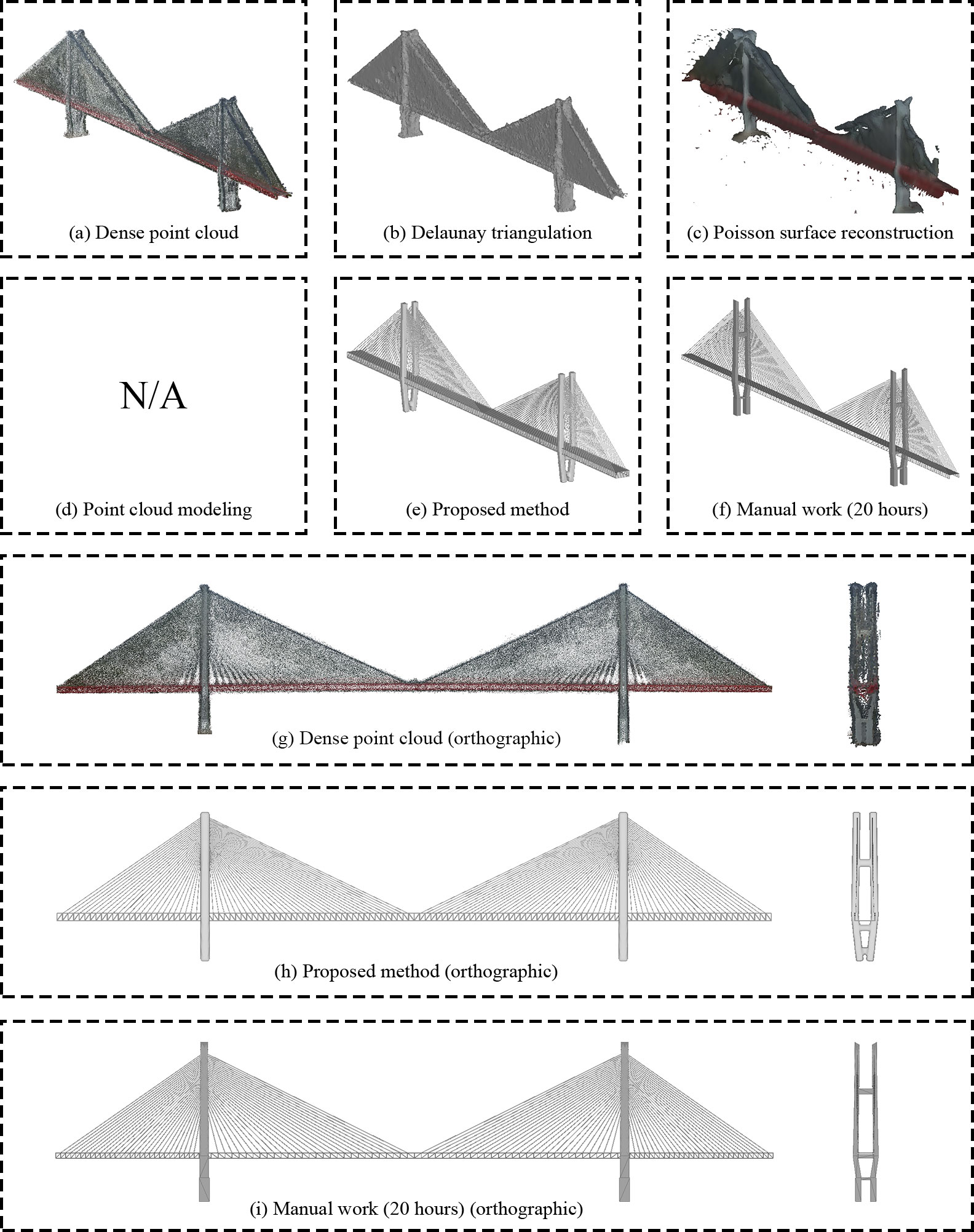}
	\caption{Bridge 1. A qualitative comparison of different methods. In addition to these methods, we manually create a 3D model based on point cloud and images as a reference.}
	\label{fig:comparison_1}  
\end{figure}

\begin{figure}[htbp]
	\centering	
	\includegraphics[width=\textwidth]{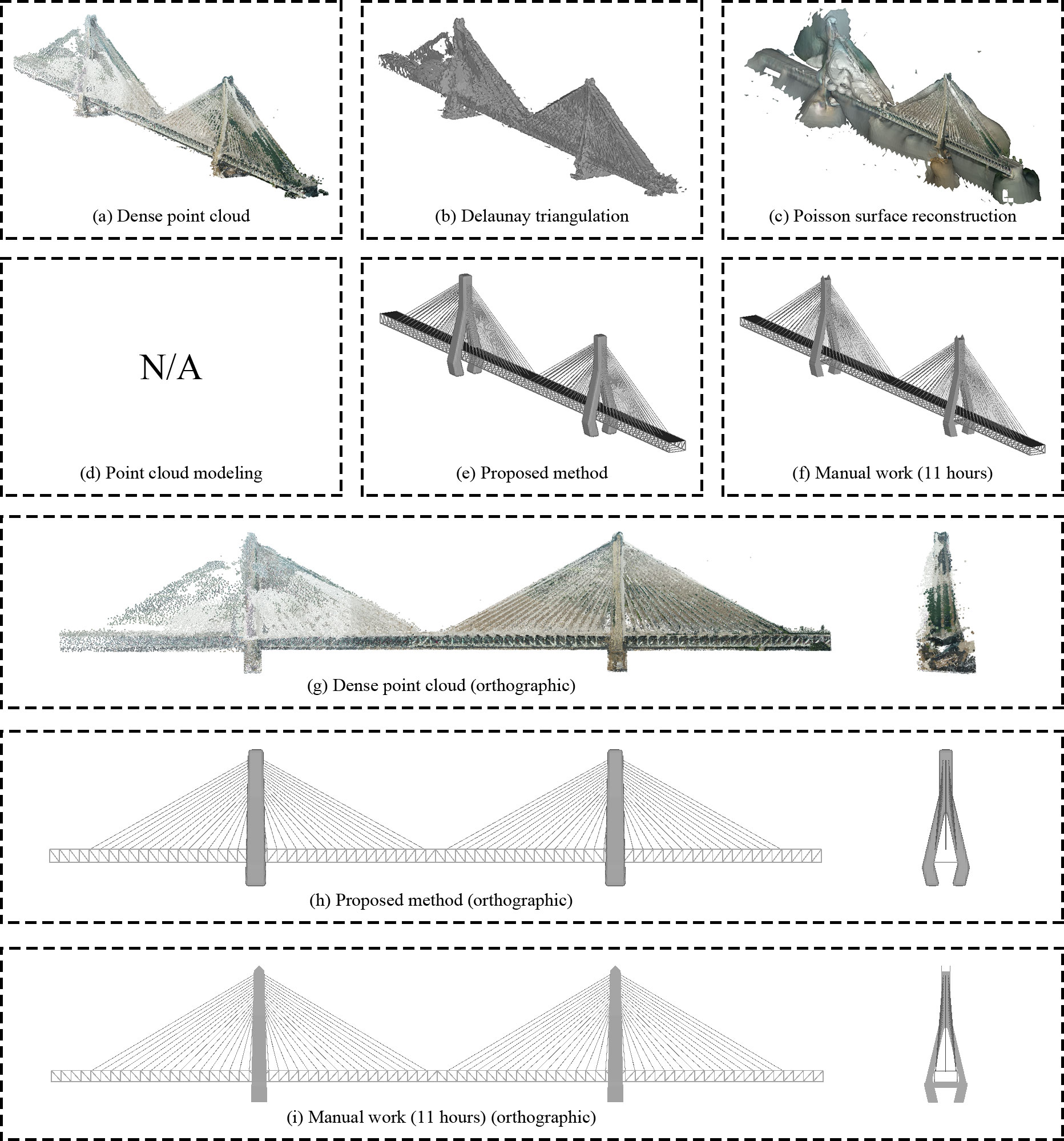}
	\caption{Bridge 2. A qualitative comparison of different methods. In addition to these methods, we manually create a 3D model based on point cloud and images as a reference.}
	\label{fig:comparison_2}  
\end{figure}

A qualitative comparison among different methods is shown in Figure \ref{fig:comparison_1} and Figure \ref{fig:comparison_2}: (a) Dense point cloud (Section \ref{sec:2_1}). Dense point cloud is not capable for texture mapping, and suffers from severe noise and uneven distribution. (b) and (c) Delaunary triangulation and Poisson surface reconstruction \cite{delaunay1934sphere, kazhdan2006poisson, kazhdan2013screened}. Surface reconstruction is directly based on point clouds and no structure prior is introduced so the meshing quality is unpleasant. (d) Point cloud modeling. This method works well in practice for high quality point clouds, we have tested on RANSAC based methods \cite{schnabel2007efficient} and Learning-based method \cite{li2018supervised}, both failed to fit 3D primitives in this case, this is due to the severe noise on MVS point cloud caused by long distance photography and limited image quantities. (e) The proposed method. (f) Manual work. (g$\sim$h) Front and side views of selected methods.

We haven’t tested image-based learning methods, since that mesh-based learning \cite{jack2018learning, wang2018pixel2mesh, groueix2018papier, kanazawa2018learning, smith2018multi} haven’t solve the topological genus problem; current geometric primitives-based methods \cite{tulsiani2017learning, zou20173d, niu2018im2struct} only take cuboids into consideration; volumetric methods \cite{choy20163d, yan2016perspective, kar2017learning, sun2018pix3d} are limited by voxel resolution. More importantly, to our best knowledge, no 3D priors are employed since no 3D information or stereo vision mechanisms are adopted in their networks. As the question \textit{"What Do Single-view 3D Reconstruction Networks Learn?"} asked in \cite{tatarchenko2019single}, current state of the art in single-view object reconstruction does not actually perform reconstruction but image classification.

\subsection{Quantitative comparison}
We use manually created 3D model as the ground truth (GT), although it seems not perform best in spatial accuracy. Chamfer distance is calculated with vertices of manually created 3D model and other models. Since Chamfer distance requires a huge KD Tree to handle a large number of vertices, hence we down-sampled to under 10000 vertices. For point cloud, the vertices are down-sampled 3D points. We also compared the reconstructed bridge components number, note that the “components” concept does not exist in the first three methods. Finally, we compared the output data size, one point has a size of 3, one triangular face is 3, one cuboid is 8, and one cylinder is 7. See Table \ref{tab:evaluation} for a comparison.

\begin{table}
	\centering
	\caption{A quantitative comparison of different methods.}
	\label{tab:evaluation}
	\setlength\heavyrulewidth{0.35ex}
	\begin{tabular}{lllll}
		\toprule
		& Method                                                               & Chamfer distance & No. of components    & Data size   \\
		\midrule
		\multirow{5}{*}{Bridge 1} 
		& Dense point cloud \cite{schonberger2016structure}                    & 0.04418          & -                    & 1117761   \\
		& Delaunay triangulation \cite{delaunay1934sphere}                     & 0.04418          & -                    & 14031345  \\
		& Poisson surface reconstruction \cite{kazhdan2013screened}            & 0.55827          & -                    & 1234419   \\
		& Point cloud modeling \cite{schnabel2007efficient, li2018supervised}  & -                & -                    & -           \\
		& The proposed method                                                  & 0.06237          & \textbf{1488}        & \textbf{66812}      \\
		& Manual work (GT)                                                     & 0                & 1430                 & 15872      \\
		\midrule
		\multirow{5}{*}{Bridge 2} 
		& Dense point cloud \cite{schonberger2016structure}                    & 0.11345          & -                    & 5398473   \\
		& Delaunay triangulation \cite{delaunay1934sphere}                     & 0.11345          & -                    & 29836674+   \\
		& Poisson surface reconstruction \cite{kazhdan2013screened}            & 0.19662          & -                    & 4307535   \\
		& Point cloud modeling \cite{schnabel2007efficient, li2018supervised}  & -                & -                    & -           \\
		& The proposed method                                                  & \textbf{0.02001} & \textbf{914}         & \textbf{73016}      \\
		& Manual work (GT)                                                     & 0                & 914                  & 10952      \\
		\bottomrule
	\end{tabular}
\end{table}

\section{CONCLUSIONS}
A learning framework is designed which can learn a mathematical model from prior knowledge. Compared with previous methods, the proposed method reconstructs a 3D model while preserving its topological properties and spatial relations successfully. 

The 3D digital model shows several potential applications, including: (a) Visual structural health monitoring. By assembling 2D images into a 3D digital bridge model using texture mapping, it enables inspectors to engage with these images in a more intuitive manner, like VR and AR. It also provides a better monitoring scheme by recording and visualizing the life cycle of an entire bridge. (b) Finite element modeling. The proposed method provides a topology-aware 3D digital model, which reduce the time spending on manual establishment of finite element model significantly.

\section{ACKNOWLEDGEMENTS}
The study is financially supported by the National Natural Science Foundation of China (Grant U1711265 and 51638007) and supported by grants from the National Key R\&D Program of China (Grant 2017YFC1500603).

%\clearpage 
\bibliographystyle{ieeetr}
\bibliography{reference}

\end{document}